\documentclass[11pt,a4paper]{article}
\usepackage[hyperref]{emnlp2020}
\usepackage{times}
\usepackage{latexsym}

\usepackage{microtype}

\aclfinalcopy %
\def\aclpaperid{2370} %

\usepackage{mystyle}

\title{Word Rotator's Distance}

\author{%
Sho Yokoi${}^{\,1,2}$
\quad\hspace{-5pt}
Ryo Takahashi${}^{\,1,2}$
\quad\hspace{-5pt}
Reina Akama${}^{\,1,2}$
\quad\hspace{-5pt}
Jun Suzuki${}^{\,1, 2}$
\quad\hspace{-5pt}
Kentaro Inui${}^{\,1, 2}$
\\
${}^{1}$ Tohoku University \quad
${}^{2}$ RIKEN\\
{\tt \{yokoi,\hspace{0.2em}ryo.t,\hspace{0.2em}reina.a,\hspace{0.2em}jun.suzuki,\hspace{0.2em}inui\}@ecei.tohoku.ac.jp}
}
\date{}

\begin{document}
\maketitle
\setcounter{page}{2944}

\begin{abstract}
A key principle in assessing textual similarity is measuring the degree of semantic overlap between two texts by considering the word alignment.
Such alignment-based approaches are intuitive and interpretable; however, they are empirically inferior to the simple cosine similarity between general-purpose sentence vectors.
To address this issue, we focus on and demonstrate the fact that the \emph{norm} of word vectors is a good proxy for word importance, and their \emph{angle} is a good proxy for word similarity. Alignment-based approaches do not distinguish them, whereas sentence-vector approaches automatically use the norm as the word importance.
Accordingly, we propose a method that first decouples word vectors into their norm and direction, and then computes alignment-based similarity using earth mover's distance (i.e., optimal transport cost), which we refer to as \emph{word rotator's distance.}
Besides, we find how to ``grow'' the norm and direction of word vectors (\emph{vector converter}), which is a new systematic approach derived from sentence-vector estimation methods.
On several textual similarity datasets, the combination of these simple proposed methods outperformed not only alignment-based approaches but also strong baselines.
\footnote{The source code is available at \url{https://github.com/eumesy/wrd}}
\end{abstract}

\section{Introduction}
\label{sec:introduction}

This paper addresses the semantic textual similarity (STS) task, the goal of which is to measure the degree of semantic equivalence between two sentences~\cite{agirre2012semeval}.
High-quality STS methods can be used to upgrade the loss functions and automatic evaluation metrics of text generation tasks because a requirement of these metrics is precisely the calculation of STS~\cite{wieting2019aclbeyondblue,zhao2019emnlp:moverscore,zhang2019bertscore}.

There are two major approaches to tackling STS.
One is to measure the degree of semantic overlap between texts by considering the word alignment, which we refer to as \emph{alignment-based approaches}~\MaybeCanOmit{\cite{sultan2014dls,pmlr-v37-kusnerb15:from,zhao2019emnlp:moverscore}}.
The other approach involves generating general-purpose sentence vectors from two texts (typically comprising word vectors), and then calculating their similarity, which we refer to as \emph{sentence-vector approaches}~\MaybeCanOmit{\cite{arora2017simplebut:sif,ethayarajh:2018:W18-30:uSIF}}.
Alignment-based approaches are consistent with human intuition about textual similarity, and their predictions are interpretable. However, the performance of such approaches is lower than that of sentence-vector approaches.

We hypothesize that one reason for the inferiority of alignment-based approaches is that they do not separate the \emph{norm} and \emph{direction} of the word vectors.
In contrast, sentence-vector approaches automatically exploit the norm of the word vectors as the relative importance of words.

Thus, we propose an STS method that first decouples word vectors into their norms and direction vectors and then aligns the direction vectors using earth mover's distance (EMD).
Here, the key idea is to map the \emph{norm} and \emph{angle} of the word vectors to the EMD parameters \emph{probability mass} and \emph{transportation cost}, respectively.
The proposed method is natural from both optimal transport and word embeddings perspectives, preserves the features of alignment-based methods, and can directly incorporate sentence-vector estimation methods, which results in fairly high performance.

Our primary contributions are as follows.
\begin{itemize}
    \item  We demonstrate that the norm of a word vector implicitly encodes the importance weight of a word and that the angle between word vectors is a good proxy for the dissimilarity of words.
    \item We propose a new textual similarity measure, i.e., word rotator's distance, that separately utilizes the norm and direction of word vectors.
    \item To enhance the proposed WRD, we utilize a new word-vector conversion mechanism, which is formally induced from recent sentence-vector estimation methods.
    \item We demonstrate that the proposed methods achieve high performance compared to strong baseline methods on several STS tasks.
\end{itemize}

\section{Task and Notation}
\label{sec:STS}

\emph{Semantic textual similarity (STS)} is the task of measuring the degree of semantic equivalence between two sentences \citep{agirre2012semeval}.
For example, the sentences
``\textit{Two boys on a couch are playing video games.}'' and ``\textit{Two boys are playing a video game.}'' are mostly equivalent (the similarity score of~$4$ out of~$5$) while the sentences ``\textit{The woman is playing the violin.}'' and ``\textit{The young lady enjoys listening to the guitar.}'' are not equivalent but on the same topic (score of~$1$)~\cite{agirre2013starsem}.
System predictions are customarily evaluated by Pearson correlation with the gold scores.
Hence, systems are only required to predict relative similarity rather than absolute scores.

We focus on \emph{unsupervised} English STS, following~\citet{arora2017simplebut:sif} and \citet{ethayarajh:2018:W18-30:uSIF}.
That is, we utilize only pre-trained word vectors, 
and do not use any supervision including
training data for related tasks (e.g., natural language inference)
and external resources (e.g., paraphrase database).
Note that \emph{semi-supervised} methods that utilize such external corpora have also been successful in English STS. 
However, the need for external corpora is a major obstacle when applying STS, a fundamental technology, to low-resource languages.
Formally, given sentences $s$ and $s'$ consisting of $n$ and $n'$ words from the vocabulary $\mathcal V$
\begin{align}
    s = (w^{}_1,\dots,w^{}_n),\;
    s' = (w'_1,\dots,w'_{n'})
    \text{,}
\end{align}
the goal is to predict the similarity $\mathrm{sim}(s,s')\in\mathbb R$.
Bold face $\VEC w^{}_i \in \mathbb R^d$ denotes the word vector corresponding to word $w^{}_i$.
Let $\langle\cdot,\cdot\rangle$ and $\norm{\cdot}$ denote the dot product and the Euclidean norm, respectively
\begin{align}
    \langle \VEC w, \VEC w' \rangle := \VEC w^\top \VEC w',\quad \norm{\VEC w} := \sqrt{\langle \VEC w, \VEC w\rangle}
    \text{.}
\end{align}

\section{Related Work} \label{sec:related-work}
\MaybeCanOmit{%
}%
We briefly review the methods that are directly related to unsupervised STS. 
\paragraph{Alignment-based Approach.}
One major approach for unsupervised STS is to compute the degree of semantic overlap between two texts~\cite{sultan2014dls,sultan2015dls}.
Recently, determining the soft alignment between word vector sets has become a mainstream method.
Tools used for alignment include attention mechanism~\cite{zhang2019bertscore}, fuzzy set~\cite{zhelezniak2019:goforthemax:fuzzyjaccard}, and earth mover's distance (EMD)~\cite{pmlr-v37-kusnerb15:from,clark2019acl:sentencemoverssimilarity,zhao2019emnlp:moverscore}.

Of those, EMD has several unique advantages.
First, it has a rich theoretical foundation for measuring the differences between probability distributions in a metric space~\cite{villani2009ot,peyre2019ot}. 
Second, EMD can incorporate structural information such as syntax trees~\cite{alvarezmelis2018aistats:structuredoptimaltransport,titouan2019icml:optimaltransportforstructureddata}.
Finally, with a simple modification, EMD can be differentiable 
and can be incorporated into larger neural networks~\cite{cuturi2013sinkhorn}.
Despite these advantages, EMD-based methods have underperformed sentence-vector-based methods on STS tasks.
The goal of this study is to identify and resolve the obstacles faced by EMD-based methods~(Section~\ref{sec:word_rorators_distance}).

\paragraph{Sentence-vector Approach.}
Another popular approach is to employ general-purpose sentence vectors of given texts and to compute the cosine similarity between such vectors.
A variety of methods to compute sentence vectors have been proposed, ranging from
utilizing deep sentence encoders~\cite{kiros2015:skip-thought,conneau2017infersent,cer2018use},
learning and using word vectors optimized for summation~\cite{pagliardini2018sent2vec,wieting2018paranmt},
and estimating latent sentence vectors from pre-trained word vectors~\cite{arora2017simplebut:sif,ethayarajh:2018:W18-30:uSIF,liu2019AAAIconceptor}.
This paper demonstrates that some recently proposed sentence vectors can be reformulated as a sum of the converted word vectors. By utilizing the converted word vectors, our method can achieve similar or better performance compared to sentence-vector approaches~(Section~\ref{sec:vector_converter}).

\section{Word Mover's Distance and its Issues}
\begin{figure}[tb]
    \centering
    \begin{tabular}{c}
        \begin{minipage}[t][4cm][b]{0.48\hsize}
            \centering
            \includegraphics[width=0.95\columnwidth, keepaspectratio=true]{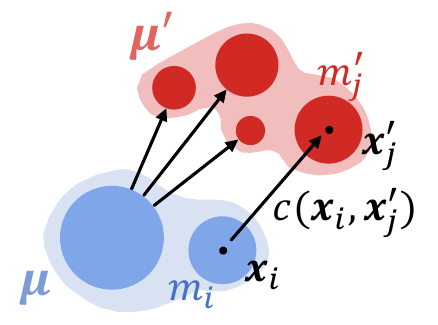}
            \vspace{4mm}
            \caption{\\Earth Mover's Distance.}
            \label{fig:emd}
        \end{minipage}
        \begin{minipage}[t][4cm][b]{0.48\hsize}
            \centering
            \includegraphics[width=1\columnwidth, keepaspectratio=true]{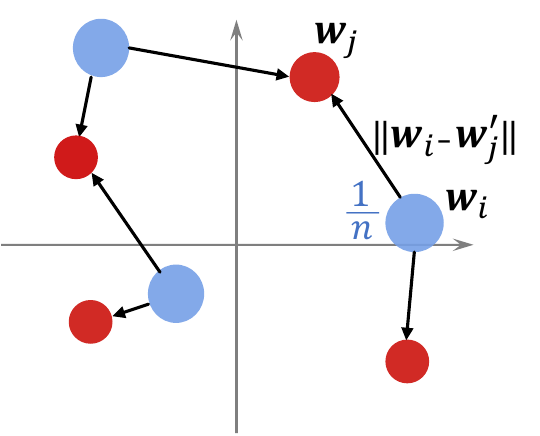}
            \caption{\\Word Mover's Distance.}
            \label{fig:wmd}
        \end{minipage}
    \end{tabular}
\end{figure}

\MaybeCanOmit{%
}
\subsection{Earth Mover's Distance}
Intuitively, \emph{earth mover's distance (EMD)}\footnote{
In this paper, following convention, we use the term earth mover's distance in the sense of optimal transport cost according to the Kantrovich formulation.
If the cost is a distance, it can also be called the $1$-Wasserstein distance.
}~\cite{villani2009ot,santambrogio2015optimaltransportforappliedmathematics,peyre2019ot}
is the minimum cost required to turn one pile of dirt into another pile of dirt (Figure~\ref{fig:emd}).
Formally, EMD takes the following inputs.
\begin{mdframed}[
    backgroundcolor=black!5,
    topline=false, bottomline=false, rightline=false, leftline=false,
    innertopmargin=0.5em,
    innerleftmargin=0.5em,
    innerbottommargin=0.5em,
    innerrightmargin=0.5em,
]
\begin{enumerate}
    \item Two \textbf{probability distributions}, $\VEC \mu$ (initial arrangement) and $\VEC \mu'$ (final arrangement)\protect\footnotemark:
    \begin{align}
        &\VEC \mu = \Big\{\! (\VEC x^{}_i,\, m^{}_i)\!\Big\}_{i=1}^{n},
        \;
        \VEC \mu' = \Big\{\! (\VEC x'_j,\, m'_j) \!\Big\}_{j=1}^{n'}
        \label{eq:discrete_probability_measure}
        \text{.}
    \end{align}
    Here, $\VEC \mu$ denotes a probability distribution, where each point $\VEC x^{}_i\in\mathbb R^d$ has a probability mass $m^{}_i\in [0,1]$ ($\sum_i m^{}_i = 1$).
    In Figure~\ref{fig:emd}, each circle represents a pair $(\VEC x^{}_i,m^{}_i)$, where the location and size of the circle represent a vector $\VEC x^{}_i$ and its probability $m^{}_i$, respectively.
    \item The \textbf{transportation cost function}, $c$:
    \begin{align}
        &c\colon \mathbb R^d\times \mathbb R^d\to \mathbb R
        \text{.}
        \label{eq:ground_metric}
    \end{align}
    Here, $c(\VEC x^{}_i, \VEC x'_j)$ determines the transportation cost per unit amount (distance) between two points $\VEC x^{}_i$ and $\VEC x'_j$.
\end{enumerate}
\end{mdframed}
\footnotetext{
    Strictly speaking, Equation \ref{eq:discrete_probability_measure} is $\VEC \mu = \sum_{i=1}^n m^{}_i\Dirac{\VEC x^{}_i}$, where the Dirac delta function describes a discrete probability measure. In this paper, we omit delta for notational simplicity.
}

The EMD between $\VEC \mu$ and $\VEC \mu'$ is then defined via the following optimization problem:
\begingroup
\allowdisplaybreaks
\begin{align}
    & \!\!\textsc{EMD}(\VEC \mu, \VEC \mu'; c)
    := \,\min_{\mathclap{{\boldsymbol T} \in \mathbb R^{n\times n'}_{\geq 0}}}
    \;\;\,
    \sum_{i,j} \VEC T_{ij}^{}\, c(\VEC x^{}_i, \VEC x'_j)
    \text{,}
    \\
    &
    \enskip
    \text{s.t.}
    \enskip
    \begin{dcases}
        \VEC T \mathds{1}^{}_n = \VEC m := (m^{}_1,\dots,m^{}_n)^\top,\\
        \VEC T^\top \mathds{1} ^{}_{n'} = \VEC m'  := (m'_1,\dots,m'_{n'})^\top.
    \end{dcases}
\end{align}
\endgroup
Here, a solution $\VEC T\in\mathbb R_{\geq 0}^{n\times n'}$ denotes a transportation plan,
where each element $\VEC T_{ij}^{}$ represents the mass transported from $\VEC x_i$ to $\VEC x'_j$.
In summary, $\textsc{EMD}(\VEC \mu,\VEC \mu'; c)$ is the cost of the best transportation plan between two distributions $\VEC \mu$ and $\VEC \mu'$. 
\paragraph{Side Benefit: Alignment.}
Under the above optimization,
if the locations $\VEC x^{}_i$ and $\VEC x'_j$ are \emph{close} (i.e., if the transportation cost $c(\VEC x^{}_i, \VEC x'_j)$ is small), they are likely to be \emph{aligned} (i.e., $\VEC T_{ij}^{}$ may be assigned a large value).
In this way, EMD can be considered to align the points of two discrete distributions.
This is one reason we adopt EMD as a key technology in the computation of STS.

\subsection{Word Mover's Distance}

\emph{Word mover's distance (WMD)}~\citep{pmlr-v37-kusnerb15:from}
is a dissimilarity measure between texts and is a pioneering work that introduced EMD to the natural language processing (NLP) field.
This study is strongly inspired by this work. We introduce WMD prior to presenting the proposed method.

WMD is the cost of transporting a set of word vectors in an embedding space (Euclidean space) (Figure~\ref{fig:wmd}).
Formally, after removing stopwords, \citet{pmlr-v37-kusnerb15:from} regard each sentence $s$ as a uniformly weighted distribution $\VEC \mu_s^{}$ comprising word vectors (i.e., bag-of-word-vectors distribution):
\begin{align}
    \!
    \VEC \mu_s^{} \!:=\! \Big\{\!(\VEC w_i^{} ,\, {\displaystyle \frac 1 n})\!\Big\}_{i=1}^{n},\;
    \VEC \mu_{s'}^{} \!:=\! \Big\{\!(\VEC w'_j,\, {\displaystyle \frac 1 {n'}})\!\Big\}_{j=1}^{n'}
    \!\text{.}\!
    \label{eq:wmd_distribution}
\end{align}
In Figure~\ref{fig:wmd}, each circle represents each word, where the location and size of the circle represent the position vector $\VEC w^{}_i$ and its weight $\frac 1 n$, respectively.
Next, Euclidean distance is used as the transportation cost between word vectors
\begin{align}
    c^{}_{\mathrm E}(\VEC w^{}_i, \VEC w'_j) := \norm{\VEC w_i^{} \!-\! \VEC w'_j}
    \text{.}
    \label{eq:euclidean_distance}
\end{align}
Then, WMD is defined as the EMD between two such distributions using the cost function $c^{}_{\mathrm E}$%
\begin{align}
    \textrm{WMD}(s,s') := \textrm{EMD}(\VEC \mu_s^{}, \VEC \mu_{s'}^{}; c_{\mathrm{E}}^{})
    \text{.}
    \label{eq:wmd}
\end{align}

\subsection{Issues with Word Mover's Distance}
\label{sec:problem_of_wmd}
Despite its intuitive formulation, the WMD often misaligns words with each other, and the STS performance of WMD is less than that of recent methods.
For example, by WMD, ``noodle'' and ``snack'' may be aligned rather than ``noodle'' and ``pho'' (a type of Vietnamese noodle).

\section{Word Rotator's Distance}
\label{sec:word_rorators_distance}

Here, we first discuss the roles of the norm and direction of word vectors. 
Then, we describe issues with WMD from the perspective of the roles of the norm and direction.
Finally, we present the proposed method, i.e., word rotator's distance, which can resolve the issues with WMD.

\subsection{Roles of Norm and Direction}
We hypothesis that the norm and direction of word vectors have the following unique roles.
\begin{itemize}
    \item \textbf{Norm of a word vector as weighting factor}: The norm of a word vector indicates the extent to which the word contributes to the overall meaning of a sentence.
    \item \textbf{Angle between word vectors as dissimilarity}: The angle between two word vectors (i.e., the difference between the direction of these vectors) approximates the (dis)similarity of two words.
\end{itemize}
We elaborate on the validity of this hypothesis in this section.
Henceforth, $\lambda^{}_i$ and $\VEC u^{}_i$ denote the norm and the direction vector of word vector $\VEC w^{}_i$, resp.:
\begin{align}
    \lambda^{}_i := \norm{\VEC w^{}_i},
    \enspace
    \VEC u^{}_i := \VEC w^{}_i / \lambda^{}_i
    \enspace
    \bigl(\VEC w^{}_i = \lambda^{}_i \VEC u^{}_i \bigr)
    \text{.}\!
    \label{eq:norm_and_direction}
\end{align}
Each $\VEC u^{}_i$ is a unit vector $\bigl(\norm{\VEC u^{}_i}=1\bigr)$.

\paragraph{Additive Compositionality.}
As a starting point, we review the well-known nature of additive compositionality. 
The NLP community has confirmed that a simple sentence vector, i.e., the average of the vectors of the words in a sentence, can achieve remarkable results when used in STS tasks and many downstream tasks~\citep{mitchell2010composition,mikolov:2013:word2vec,Wieting2016TowardsUP}.
\begingroup
\allowdisplaybreaks
\begin{align}
    &
    \VEC s^{}_{\textsc{add}}
    =
    \frac 1 n
    \sum_{w^{}_i\in s} \VEC w^{}_i,
    \quad
    \VEC s'_{\textsc{add}}
    =
    \frac 1 {n'}
    \sum_{w'_j\in s'} \VEC w'_j,
    \label{eq:additive_composition}
    \\
    &
    \mathrm{sim}(s,s') = \cos(\VEC s^{}_{\textsc{add}}, \VEC s'_{\textsc{add}})
    \label{eq:additive_composition_cosine}
    \text{.}
\end{align}
\endgroup

\begin{figure}
    \centering
    \includegraphics[width=0.7\columnwidth, keepaspectratio=true]{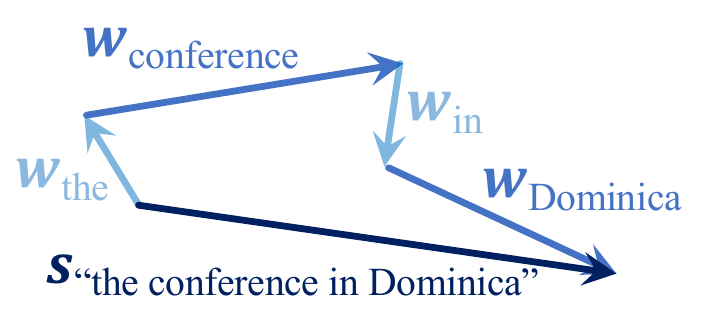}
    \caption{The addition operation implicitly uses the norm of the vectors as the weighting factor.}
    \label{fig:additive_composition}
\end{figure}

\paragraph{Norm as Weighting Factor.}
Equation~\ref{eq:additive_composition} may initially appear to treat each word vector equally.
However, several previous studies have confirmed that the norm of word vectors has large dispersion~\cite{schakelandwilson2015:measuringwordsignificance,arefyev2018:howmuchdoesawordweigh}.
In other words, a sentence vector would contain word vectors of various \Edited{lengths}.
In such cases, a word vector with a large norm will dominate in the resulting sentence vector, and vice versa~(Figure~\ref{fig:additive_composition}).
Here the usefulness of additive composition (i.e., implicit weighting by the norm) suggests that the norm of each word vector functions as the weighting factor of the word when generating a sentence representation.
In our experiments, we provide data-driven evidence to support this claim.

In addition, the following are known about the relationship between word vector norm and the word importance:
(i) content words tend to have larger norms than function words~\cite{schakelandwilson2015:measuringwordsignificance}; and
(ii) fine-tuned word vectors have larger norms for medium-frequency words, which is consistent with the traditional weighting guideline by Luhn in information retrieval~\cite{khodak-EtAl:2018:alacarte,pagliardini2018sent2vec}.
Both of these observations suggest that the norm serves as a weighting factor in cases where additive composition is effective.

\paragraph{Angle as Dissimilarity.}
What does a direction vector (i.e., the rest of the word vector ``minus'' its norm) represent?\footnote{%
    Analogous to the polar coordinate system, Equation~\ref{eq:norm_and_direction} decouples each word vector into a one-dimensional norm and a $(d-1)$-dimensional direction vector.
}
Obviously, the most common calculation using the direction vectors of words is to measure their angles, i.e., their cosine similarity
\begin{align}
    \cos(\VEC w, \VEC w')
    = \frac{\langle \VEC w, \VEC w'\rangle}{\lambda  \lambda'}    
    = \langle \VEC u, \VEC u'\rangle
    \text{.}
\end{align}
It is widely known that the cosine similarity of word vectors trained on the basis of the distributional hypothesis well approximates word similarity~\cite{pennington-socher-manning:2014:EMNLP2014:glove,mikolov:2013:word2vec,bojanowski2017fasttext}. 
Naturally, the difference in direction vectors represents the dissimilarity of words.
In our experiments, we confirmed that cosine similarity is an empirically better proxy for word similarity compared to other measures.

\subsection{Why doesn't WMD Work?}
\begin{figure}
    \centering
    \includegraphics[width=0.65\columnwidth, keepaspectratio=true]{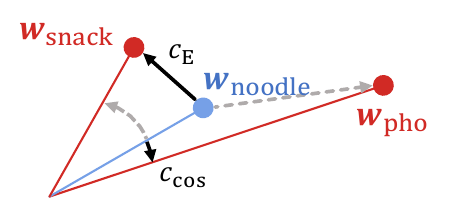}
    \caption{Euclidean distance ``mixes up'' the norm (a weighting factor for each word) and direction vectors (for word dissimilarity).}
    \label{fig:L2_issue}
\end{figure}
According to the above discussion, WMD has the following limitations.
\begin{itemize}
    \item \textbf{Weighting of words:}
    While EMD can consider the weights of each point via their probability mass~\eqref{eq:discrete_probability_measure}, and the weighting factor of each word is encoded in the norm,
    WMD ignores the norm and weights each word vector uniformly~\eqref{eq:wmd_distribution}.
    \item \textbf{Dissimilarity between words:}
    While EMD can consider the distance between points via a transportation cost~\eqref{eq:ground_metric}, and the dissimilarity between words can be measured by angle,
    WMD uses Euclidean distance, which \emph{mixes} the weighting factor and dissimilarity.
\end{itemize}

The problematic nature of this \emph{mixing} can be explained as follows.
Euclidean transportation cost~\eqref{eq:euclidean_distance} would misestimate the similarity of word pairs as low\textsubscript{$\langle$A$\rangle$},
whose meanings are close\textsubscript{$\langle$B$\rangle$} but whose concreteness or importance is very different\textsubscript{$\langle$C$\rangle$}, e.g., ``noodle'' and ``pho'' (Figure~\ref{fig:L2_issue}).
This is clear from the relationship between the Euclidean~\eqref{eq:euclidean_distance} and cosine distances~\eqref{eq:cosine_distance_for_L2}:
\begin{align}
    & c^{}_{\cos}(\VEC w, \VEC w') := 1 - \cos(\VEC w, \VEC w')
    \label{eq:cosine_distance_for_L2}
    \\
    & c^{}_{\mathrm{E}}(\VEC w, \VEC w')
    = \sqrt{(\lambda \VEC u - \lambda' \VEC u')^\top(\lambda \VEC u - \lambda' \VEC u')}\\
    & \quad = \sqrt{\lambda\lambda' \left(2 c^{}_{\cos}(\VEC w, \VEC w') + (\lambda - \lambda')^2\right)}
    \label{eq:relation_L2_cos}
    \text{.}
\end{align}
From Equation~\ref{eq:relation_L2_cos}, $c^{}_{\mathrm E}(\VEC w, \VEC w')$ would be estimated as large\textsubscript{$\langle$A$\rangle$} even if $c^{}_{\cos}(\VEC w, \VEC w')$ is small\textsubscript{$\langle$B$\rangle$},
as long as $\abs{\lambda - \lambda'}$ is large\textsubscript{$\langle$C$\rangle$}.
Note that this undesirable property is also confirmed when using real data.
Table~\ref{tb:issue_wmd} and Figure~\ref{fig:L2_issue} show the cosine and Euclidean distances between the vectors of ``noodle,'' ``pho,'' ``snack,'' and ``Pringles'' (the name of a snack).
By using Euclidean distance, ``noodle'' and ``snack'' are judged to be similar (i.e., more likely to be aligned) than ``noodle'' and ``pho.''
\begin{table}[tb]
\begin{subtable}[t]{0.56\columnwidth}
\centering
\setlength{\tabcolsep}{.4pt} %
\footnotesize
\begin{tabular}{l cccc}
\toprule
& \condensed{noodle} & \condensed{pho} & \condensed{snack} & \condensed{Pringles} \\
\midrule
\condensed{noodle} & - & \textbf{0.43} & 0.58 & 0.83 \\
\condensed{pho} & \textbf{0.43} & - & 0.73 & 0.94 \\
\condensed{snack} & 0.58 & 0.73 & - & \textbf{0.56} \\
\condensed{Pringles}\hspace{-1pt} & 0.83 & 0.94 & \textbf{0.56} & - \\
\bottomrule
\end{tabular}
\caption{Cosine distance.}
\label{tb:issue_wmd_cosine}
\end{subtable}
\begin{subtable}[t]{0.43\columnwidth}
\centering
\setlength{\tabcolsep}{.4pt} %
\footnotesize
\begin{tabular}{cccc}
\toprule
\condensed{noodle} & \condensed{pho} & \condensed{snack} & \condensed{Pringles} \\
\midrule
- & 3.52 & \uline{\textbf{3.39}} & 4.62 \\
\textbf{3.52} & - & 4.52 & 5.60 \\
\uline{\textbf{3.39}} & 4.52 & - & 3.84 \\
4.62 & 5.60 & \textbf{3.84} & - \\
\bottomrule
\end{tabular}
\caption{Euclidean distance.}
\label{tb:issue_wmd_euclidean}
\end{subtable}
\caption{%
    Differences in behavior between cosine and Euclidean distance.
    For each row, the lowest value (closest word) is shown in \textbf{bold}.
    Inappropriate alignments by Euclidean distance are \uline{\textbf{underlined}}.
    Here, pre-trained word2vec~\cite{mikolov:2013:word2vec} was used.
}
\label{tb:issue_wmd}
\end{table}%

\subsection{Word Rotator's Distance}
\begin{figure}
    \centering
    \includegraphics[width=0.7\columnwidth, keepaspectratio=true]{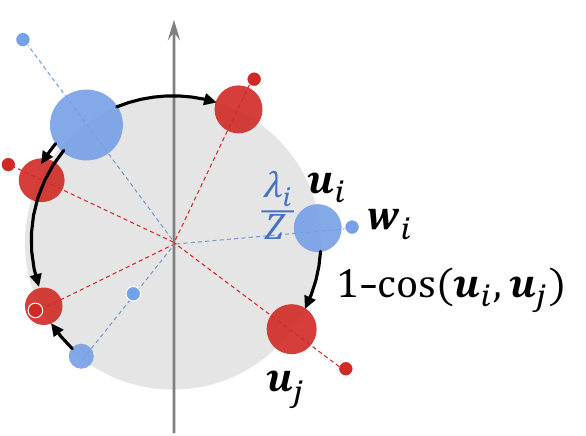}
    \caption{Word Rotator's Distance.}
    \label{fig:wrd}
\end{figure}

Given the above considerations, we propose a simple yet powerful sentence similarity measure using EMD.
The proposed method considers each sentence as a discrete distribution on the unit hypersphere and calculates EMD on this hypersphere (Figure~\ref{fig:wrd}).
Here, the alignment of the direction vectors corresponds to a rotation on the unit hypersphere; thus, we refer to the proposed method as \emph{word rotator's distance (WRD)}.

Formally, we consider each sentence $s$ as a discrete distribution $\VEC \nu^{}_s$ comprising direction vectors weighted by their norm (bag-of-direction-vectors distribution)
\begingroup
\allowdisplaybreaks
\begin{align}
    &
    \VEC \nu^{}_{s}    \!:=\! \Big\{(\VEC u^{}_i,\, \frac {\lambda^{}_i}{Z})\Big\}_{i=1}^{n},
    \;
    \VEC \nu^{}_{s'}    \!:=\! \Big\{(\VEC u'_j,\, \frac {\lambda^{}_j}{Z'})\Big\}_{j=1}^{n'}
    \text{,}
    \label{eq:sentence_dist_wrd}
\end{align}
\endgroup
where $Z$ and $Z'$ are normalizing constants
($Z:=\sum_i \lambda_i$, so as $Z'$).
In Figure~\ref{fig:wrd}, each circle represents a word, 
where the location and size of the circle represent the direction vector $\VEC u^{}_i$
and its weight $\lambda^{}_i/Z$, respectively.
For the cost function, we use the cosine distance
\begin{align}
    & c^{}_{\cos}(\VEC u^{}_i, \VEC u'_j) = 1 - \cos(\VEC u_i^{}, \VEC u'_j)
    \label{eq:cosine_distance}
\end{align}
That is, a rotation cost is required to align words.
Then, the WRD between two sentences is given as
\begin{align}
    \textrm{WRD}(s,s') := \textrm{EMD}(\VEC \nu_s^{}, \VEC \nu_{s'}^{}; c^{}_{\cos})
    \text{.}
\end{align}

Unlike WMD, the above procedure allows the proposed WRD to follow appropriate correspondences between the EMD and word vectors.
\begin{mdframed}[
    backgroundcolor=black!5,
    topline=false, bottomline=false, rightline=false, leftline=false,
    innertopmargin=0.5em,
    innerleftmargin=0.5em,
    innerbottommargin=0.5em,
    innerrightmargin=0.5em,
]
\begin{itemize}
    \item Probability mass (\textbf{weight} of each point)\\ $\leftrightarrow$ Norm (\textbf{weight} of each word)
    \item Transportation cost (\textbf{distance} between points)\\ $\leftrightarrow$ Angle (\textbf{dissimilarity} between words)
\end{itemize}
\end{mdframed}

\paragraph{Algorithm.}
To ensure reproducibility, we show the specific (and fairly simple) algorithm and implementation guidelines for WRD in Appendix~\ref{sec:appendix:algorithm_of_wrd}.
\section{Vector Converter-enhanced WRD}
\label{sec:vector_converter}

To further improve the performance of WRD, we attempted to integrate existing methods to estimate latent sentence vectors, i.e., the most powerful sentence encoders for STS, into WRD.
However, determining a method to combine sentence-vector estimation methods with WRD is not straightforward task because WRD takes \emph{word vectors} as input, whereas sentence-vector estimation methods require the processing of \emph{sentence vectors}.

\subsection{From Sentence Vector to Word Vector}
\label{sec:vector_converter_subsec}

\Edited{%
}

\paragraph{Sentence-vector Estimation.}

On the basis of Arora's pioneering random-walk language model (LM) \cite{arora2016latentvariable,arora2017simplebut:sif},
a number of sentence-vector estimation methods have been proposed~\citep{arora2017simplebut:sif,ethayarajh:2018:W18-30:uSIF,liu2019AAAIconceptor,liu2019NAACLconceptor}, and have achieved success in many NLP applications, including STS.
Given pre-trained vectors of words comprising a sentence, these methods allow us to estimate the latent sentence vectors that generated the word vectors.
Such methods can be summarized in the following form
\begin{align}
    \mathrm{Encode}(s)
    =
    f_3^{}\biggl(\frac{1}{n}\sum_{w\in s}^{} \alpha_2^{}(w) f_1^{}(\VEC w) \!\biggr)
    \text{,}
    \label{eq:affine_encoder}
\end{align}
where
\begin{itemize}
    \item 
    $f^{}_1\colon \mathbb R^D\!\to\mathbb R^D$ ``denoises'' each word vector; %
    \item 
    $\alpha^{}_2\colon\mathcal V\to\mathbb R\;$ 
    scales each word vector;
    \item 
    $f^{}_3\colon\mathbb R^D\!\to\mathbb R^D$ ``denoises'' each sentence vector.
\end{itemize}
Here, we focus on only the form of the equation for sentence-vector estimation.
For specific algorithms, refer to the experimental section and Appendix~\ref{sec:appendix:algorithm_of_vc}.

\paragraph{Word Vector Converter.}
Note that all of the existing denoising function $f_3$ is linear; thus, Equation~\ref{eq:affine_encoder} can be rewritten as
\begin{align}
    & \mathrm{Encode}(s)
    = \frac 1 n \sum_{w\in s} \widetilde{\VEC w}
    \label{eq:additive_composition_affine}
    \\
    & \widetilde{\VEC w}
    = f^{}_{\mathrm{VC}}(\VEC w)
    := f^{}_3\left(\alpha_2^{}(w) \cdot f^{}_1(\VEC w)\right)
    \text{.}
    \label{eq:word_vector_converter}
\end{align}
Here, the encoders first perform a transformation $f^{}_{\mathrm{VC}}$ on each word vector independently and then sum them up (i.e., additive composition!).
We refer to the $f^{}_{\mathrm{VC}}$ as \emph{(word) vector converter (VC)}.

\subsection{Norm and Direction}
\label{sec:norm_and_direction_by_word_vector_converter}

We believe that the vector converter
improves the norm and direction of pre-trained word vectors.

\paragraph{Norm as Weighting Factor.}
In Section~\ref{sec:word_rorators_distance}, given the success of additive composition, we proposed the use of norms to weight words.
In addition, in Section~\ref{sec:vector_converter_subsec}, we confirmed that the sentence-vector estimation methods, which have achieved greater success in STS than standard additive composition, simply sum the transformed word vectors (i.e., improved additive composition).
Therefore, we expect that the importance of a word $w$ is better encoded in the norm of a converted word vector $\widetilde{\VEC w}$ than in that of the original word vector $\VEC w$.

\paragraph{Angle as Dissimilarity.}
On the bases of the random-walk LM~\cite{arora2016latentvariable}, the denoising function $f^{}_1$ makes the word vector space isotropic, i.e., uniform in the sense of angle.
As a result, the angle of word vectors becomes a better proxy for word dissimilarity~\cite{mu2018allbutthetop}.
Further, the functions $\alpha^{}_2$ and $f^{}_3$ assume a more realistic LM
\cite{arora2017simplebut:sif}.
\Edited{%
Thus, VC is expected to further improve the isotropy of the vector space and make the angle of the word vector a better proxy for word dissimilarity.}

\subsection{Vector Converter-enhanced WRD}
As we discussed previously, converted word vectors $\{\widetilde{\VEC w}\}$ may have preferable properties in terms of their norm and direction, and they remain word vectors (i.e., they are no longer sentence vectors); thus, $\{\widetilde{\VEC w}\}$ can be used as is for the input of WRD.
Let $\widetilde{\lambda}$ and $\widetilde{\VEC u}$ denote the norm and direction vector of $\widetilde{\VEC w}$, resp.; then, a variant of WRD using $\{\widetilde{\VEC w}\}$ is
\begin{align}
    & 
    \widetilde{\VEC \nu}^{}_{s}
    \!:=\! \Big\{(\widetilde{\VEC u}^{}_i,\, \frac {\widetilde{\lambda}^{}_i}{Z})\Big\}_{i=1}^{n}
    \text{,}
    \;
    \widetilde{\VEC \nu}^{}_{s'}
    \!:=\! \Big\{(\widetilde{\VEC u}'_j,\, \frac {\widetilde{\lambda}^{}_j}{Z'})\Big\}_{j=1}^{n'}
    \text{,}
    \\
    & \textrm{WRD}_{\textrm{with VC}}^{}(s,s') := \textrm{EMD}(\widetilde{\VEC \nu}^{}_s, \widetilde{\VEC \nu}^{}_{s'}; c^{}_{\cos})
    \text{,}
\end{align}
where $Z$ and $Z'$ are normalizing constants.
We believe that using $\{\widetilde{\VEC w}\}$ will improve WRD performance because WRD depends on the weights and dissimilarities encoded in the norm and angle.

\section{Experiments}
In this section, we experimentally confirm our hypotheses about the norm and direction vectors and the performance of WRD and VC.
\MaybeCanOmit{%
}
For word vectors, we used the standard \textbf{GloVe} \cite{pennington-socher-manning:2014:EMNLP2014:glove}, \textbf{word2vec} \cite{mikolov:2013:word2vec}, and \textbf{fastText} \cite{bojanowski2017fasttext}.
We did not use BERT~\cite{devlin2018bert} and its variants because they are currently ineffective in unsupervised STS tasks. Refer to the Appendix~\ref{sec:bert_for_unsupervised_sts} for additional details).

For STS datasets, we used \textbf{STS'15} \cite{agirre2015semeval} for comparison with \citet{zhelezniak2019:goforthemax:fuzzyjaccard,kiros2015:skip-thought,peters2018elmo}, STS benchmark (\textbf{STS-B}) \cite{cer2017semeval:stsb}, one of the most actively used datasets, and \textbf{Twitter'15} \cite{xu2015semeval} to validate methods against casual writing.
\begin{table*}[tb]
\small
\setlength{\tabcolsep}{1.2pt}  %
\renewcommand{\arraystretch}{1.15} %
\begin{subtable}[t]{0.40\columnwidth}
    \centering
    \begin{tabular}[t]{lcc}
    \toprule
     & \textsc{add} & \makecell{\textsc{w/o} \\ \textsc{norm}} \\
    \midrule
    GloVe & \textbf{54.16} & 46.25 \\
    word2vec & \textbf{72.43} & 63.20 \\
    fastText & \textbf{70.40} & 56.31 \\
    \bottomrule
    \end{tabular}
    \caption{Pre-trained word \\ vectors.}
    \label{subtb:ignoring_norm}
\end{subtable}
\begin{subtable}[t]{1.7\columnwidth}
    \centering
    \begin{tabular}[t]{lcccclcccclccc}
    \addlinespace[-\aboverulesep] 
    \cmidrule[\heavyrulewidth]{1-4}
    \cmidrule[\heavyrulewidth]{6-9}
    \cmidrule[\heavyrulewidth]{11-14}
     & \textsc{add} & \makecell{\textsc{w/o} \\ \textsc{norm}} & diff
     &&& \textsc{add} & \makecell{\textsc{w/o} \\ \textsc{norm}} & diff
     &&& \textsc{add} & \makecell{\textsc{w/o} \\ \textsc{norm}} & diff
    \\
    \cmidrule{1-4}
    \cmidrule{6-9}
    \cmidrule{11-14}
    GloVe & \textbf{57.60} & 50.83 & 6.77
    && word2vec & \textbf{72.22} & 62.97 & 9.25
    && fastText & \textbf{71.49} & 59.25 & 12.24
    \\
    + A & \textbf{67.56} & 59.03 & 8.53
    && + A & \textbf{72.32} & 63.37 & 8.95
    && + A & \textbf{69.67} & 57.34 & 12.33
    \\
    + AW & \textbf{76.06} & 59.03 & 17.03
    && + AW & \textbf{76.61} & 63.37 & 13.24
    && + AW & \textbf{79.09} & 57.34 & 21.75
    \\
    + \textbf{VC}(AWR) & \textbf{77.51} & 59.71 & 17.80
    && + \textbf{VC}(AWR) & \textbf{77.33} & 63.96 & 13.37
    && + \textbf{VC}(AWR) & \textbf{79.68} & 57.24 & 22.44
    \\
    \cmidrule[\heavyrulewidth]{1-4}
    \cmidrule[\heavyrulewidth]{6-9}
    \cmidrule[\heavyrulewidth]{11-14}
    \addlinespace[-\belowrulesep]
    \end{tabular}
    \caption{Converted word vectors.}
    \label{subtb:ignoring_norm_converted}
\end{subtable}

\caption{%
    Norm has a large impact on additive composition.
    Pearson's $r \times 100$ between the predicted and gold scores is reported.
    The best result in each row is indicated in \textbf{bold}.
    The STS-B dataset (dev) was used.}
\end{table*}
For VC, we used the followings algorithms.
\begin{itemize}
    \item $f_1^{}$:\!
        \textbf{A}ll-but-the-top~\citep{mu2018allbutthetop},
        sentence-wise feature \textbf{S}caling \!\cite{ethayarajh:2018:W18-30:uSIF}\footnote{%
            \citet{ethayarajh:2018:W18-30:uSIF} proposed three methods, i.e., S, U, and P. 
            For the correctness, we abbreviate this series of methods as SUP, which was abbreviated as UP in the original paper.
        }\!.\!\!
    \item $\alpha_2^{}$:
        SIF \textbf{W}eighting~\citep{arora2017simplebut:sif},
        \textbf{U}nsupervised SIF weighting~\citep{ethayarajh:2018:W18-30:uSIF}.
    \item $f_3^{}$:
        Common component \textbf{R}emoval \citep{arora2017simplebut:sif},
        \textbf{P}iecewise CCR \citep{ethayarajh:2018:W18-30:uSIF},
        \textbf{C}onceptor removal \citep{liu2019AAAIconceptor}.
\end{itemize}
Henceforth, a bold character denotes each method.
In addition, \textbf{VC}(AWR), for example, denotes \textbf{VC} induced by \textbf{A}, \textbf{W}, and \textbf{R}.
Note that we did not tune hyperparameters; we used values reported in previous studies.
See Appendix~\ref{sec:appendix:algorithm_of_vc} for details.

\subsection{Workings of Norm}
Here, we experimentally confirm whether the norm of a word vector is in fact a good proxy of the word's in-sentence importance.

\paragraph{Pre-trained Word Vectors.}
Let us consider another additive composition than that in Equation~\ref{eq:additive_composition}, which excludes the effect of weighting by the norm
\begin{align}
    \VEC s^{}_{\textsc{add w/o norm}}
    =
    \sum_{w^{}_i\in s} \VEC w^{}_i / \lambda^{}_i
    =
    \sum_{w^{}_i\in s} \VEC u^{}_i
    \text{.}
    \label{eq:additive_composition_ignoring_norm}
\end{align}
Table~\ref{subtb:ignoring_norm} shows the experimental results obtained
using two types of sentence vectors \eqref{eq:additive_composition}, \eqref{eq:additive_composition_ignoring_norm}.
Ignoring the norm of the word vectors produced consistently poor performance.
This demonstrates that the norm of a word vector certainly plays the role of the weighting factor of the word.

\paragraph{Converted Word Vectors.}
To verify our hypothesis that VC improves the norm, we performed the same experiments as above using converted word vectors.
Table~\ref{subtb:ignoring_norm_converted} shows that as the word vectors are gradually converted, the difference in predictive performance between Equation \ref{eq:additive_composition} and \ref{eq:additive_composition_ignoring_norm} (i.e., the performance gain by norm) increased.
This fact supports our hypothesis that VC ``grows'' the norm.
\MaybeCanOmit{%
}

\begin{table}[tb]
\centering
\small
\setlength{\tabcolsep}{3pt}  %
\renewcommand{\arraystretch}{1.15} %
\newcommand{\tblpaddummy}{\hspace{4pt}}
\begin{subtable}[t]{.48\linewidth}
    \centering
    \begin{tabular}[t]{lccc}
    \toprule
    & \textsc{L2} & \textsc{dot} & \textsc{cos} \\
    \vphantom{\makecell{+ \textbf{VC} \\ (AWR)}} &&& \\
    \midrule
    \multicolumn{4}{l}{GloVe} \\
    \tblpaddummy MEN       & 73.36 & \textbf{80.79} & \textbf{80.49} \\
    \tblpaddummy RW        & 45.13 & 48.17          & \textbf{51.04} \\
    \midrule
    \multicolumn{4}{l}{word2vec} \\
    \tblpaddummy MEN       & 62.31 & 74.46 & \textbf{78.20} \\
    \tblpaddummy RW        & 35.10 & 51.29 & \textbf{55.80} \\
    \midrule
    \multicolumn{4}{l}{fastText} \\
    \tblpaddummy MEN       & 68.67 & 79.37 & \textbf{84.55} \\
    \tblpaddummy RW        & 47.24 & 53.83 & \textbf{62.17} \\
    \bottomrule
    \end{tabular}
    \caption{It is better to use only direction instead of the norm.}
    \label{subtb:workings_of_angle_cos}
\end{subtable}
\hspace{0.03\linewidth}
\begin{subtable}[t]{.47\linewidth}
    \centering
    \begin{tabular}[t]{ccc}
    \toprule
    \textsc{cos} & \textsc{cos} & \textsc{cos} \\
    \makecell{original \\ vector} & + A & \makecell{+ \textbf{VC} \\ (AWR)} \\
    \midrule
    \multicolumn{3}{l}{GloVe} \\
    \tblpaddummy 80.49          & 82.58 & \textbf{83.69} \\
    \tblpaddummy 51.04          & \textbf{57.87} & \textbf{57.96} \\
    \midrule
    \multicolumn{3}{l}{word2vec} \\
    \tblpaddummy 78.20 & \textbf{80.22} & \textbf{80.15} \\
    \tblpaddummy 55.80 & \textbf{57.39} & \textbf{57.47} \\
    \midrule
    \multicolumn{3}{l}{fastText} \\
    \tblpaddummy 84.55 & \textbf{85.51} & \textbf{85.91} \\
    \tblpaddummy 62.17 & \textbf{62.98} & \textbf{62.75} \\
    \bottomrule
    \end{tabular}
    \caption{%
    VC gradually ``grows'' the direction of word vectors.}
    \label{subtb:workings_of_angle_vc}
\end{subtable}
\caption{%
    The angle of word vectors is a good proxy for word similarity.
    Spearman's $\rho$ $\times$ $100$
    between the predicted and gold scores is reported.
    In each row, the best result and results where the difference from the best result was $<0.5$ are indicated in \textbf{bold}.
    \protect\footnotemark
    }
\end{table}
\footnotetext{%
    ``+ AW'' is omitted from Table~\ref{subtb:workings_of_angle_vc} because \textsc{W} (i.e., the scaling function) alone does not change the angle.
}
\subsection{Workings of Angle}
We assumed the angle between two word vectors
is a good proxy for the dissimilarity of two words.
Presently, the cosine similarity between word vectors is a common metrics to compute word dissimilarity;
however, several alignment-based STS methods employ Euclidean distance~\cite{pmlr-v37-kusnerb15:from} or dot product~\cite{zhelezniak2019:goforthemax:fuzzyjaccard}.
Therefore, a question arises, i.e., which is the most suitable method to compute word dissimilarity?
To answer this question, we compared dissimilarity metrics using nine word similarity datasets\footnote{%
    See Appendix~\ref{sec:resources_experiments} for details regarding the datasets.
    While the results for the two larger datasets are presented here, the results for all nine datasets are presented in Appendix~\ref{sec:full_results}.
}.
\paragraph{Pre-trained Word Vectors.}
Table \ref{subtb:workings_of_angle_cos} shows that cosine similarity (i.e., ignoring the norm) yields relatively higher correlation with human evaluations compared to dot product or Euclidean distance (i.e., using the norm).
This indicates that the angle of word vectors encodes the dissimilarity of words relatively well; in contrast, the norm is not relevant.

\paragraph{Converted Word Vectors.}
In view of the discussion given in Section~\ref{sec:vector_converter}, we expected that the word dissimilarity of $w$ and $w'$ would be better encoded in the angle between the converted word vectors $\langle \widetilde{\VEC u}, \widetilde{\VEC u}'\rangle = \cos(\widetilde{\VEC w}, \widetilde{\VEC w}')$ than that between the original word vectors $\langle \VEC u, \VEC u'\rangle = \cos(\VEC w,\VEC w')$.
Table~\ref{subtb:workings_of_angle_vc} shows that, as the word vectors were gradually converted, the angle of word vectors became more accurate as a measure of the dissimilarity of words.
\subsection{Ablation Study}
We experimentally confirmed the effectiveness of each WRD and VC, through the degree of performance improvement over the baseline WMD.
Table~\ref{tb:ablation_new2} shows the results. In nearly all cases, WRD demonstrated higher performance than WMD. We summarize some major findings as follows.

\begin{table}[tb]
\centering
\setlength{\tabcolsep}{1.7pt}  %
\renewcommand{\arraystretch}{1.15} %
\small
\begin{tabular}{lcccccccc}
\addlinespace[-\aboverulesep] 
\cmidrule[\heavyrulewidth]{1-3}
\cmidrule[\heavyrulewidth]{5-6}
\cmidrule[\heavyrulewidth]{8-9}
 & \multicolumn{2}{c}{GloVe}
 && \multicolumn{2}{c}{word2vec}
 && \multicolumn{2}{c}{fastText} \\
 & WMD & \textbf{WRD}
 && WMD & \textbf{WRD}
 && WMD & \textbf{WRD} \\ 
\cmidrule{1-3}
\cmidrule{5-6}
\cmidrule{8-9}
original vector & 65.74 & \textbf{67.70} && 67.21 & \textbf{70.91} &&  64.06 & \textbf{69.31} \\
+ A & 65.44 & \textbf{68.26} && 67.09 & \textbf{71.23} && 63.79 & \textbf{69.34} \\
+ AW & 64.76 & \textbf{76.08} && 64.97 & \textbf{75.00} && 62.47 & \textbf{76.90} \\
+ \textbf{VC}(AWR) & 64.74 & \uline{\textbf{76.87}} && 64.89 & \uline{\textbf{76.04}} && 62.47 & \uline{\textbf{77.56}} \\
+ SIF weights & 75.42 & - && 73.90 & - && 74.64 & - \\
\cmidrule[\heavyrulewidth]{1-3}
\cmidrule[\heavyrulewidth]{5-6}
\cmidrule[\heavyrulewidth]{8-9}
\addlinespace[-\belowrulesep]
\end{tabular}

\caption{%
    The combination of WRD and VC gave the best performance.
    Pearson's $r \times 100$ between the predicted and gold scores is reported.
    The STS-B dataset (dev) was used.
    The best result and results where the difference from the best $<0.5$ in each row are in \textbf{bold},
    and the best results
    are further \uline{\textbf{underlined}}.
}
\label{tb:ablation_new2}
\end{table}
\begin{itemize}

    \item 
    The performance of WRD improves steadily, as word vectors are transformed by VC, because WRD can directly utilize the weight and dissimilarity encoded in the norm and angle, whose quality is enhanced by VC.
    Conversely, WMD does not benefit from VC.

    \item 
    One may consider that \textsc{W} (SIF weighting) can be used directly as the probability mass for WMD because it is simply a scaling factor for each word.
    ``+ SIF weights'' in Table~\ref{tb:ablation_new2} represents such a computation; however, even when WMD
    and employed SIF directly,
    it did not reach the performance of WRD and VC.
\end{itemize}
Following~\citet{pmlr-v37-kusnerb15:from}, we further experimented with stopword removal.
Stopword removal was a good heuristic that gave both WMD and WRD a large performance gain similar to SIF; however, the above two findings remained unchanged.
See Appendix~\ref{sec:full_results} for additional details.

\subsection{Benchmark Tasks}
\begin{table}[tb]
\centering
\bgroup
\setlength{\tabcolsep}{1pt}
\renewcommand{\arraystretch}{0.94}
\footnotesize
\begin{tabular}{@{}l ccccc ccccc@{}}
\toprule
& STS'15 & STS-B & Twitter \\
\midrule
\multicolumn{1}{l}{GloVe -- Additive Composition} \\
\tblpad GloVe$^\dagger$ & 56.08 & 45.57 & 29.35 \\
\tblpad GloVe + WR$^\dagger$ \citep{arora2017simplebut:sif} & 67.74 & 62.85 & 40.03 \\
\tblpad GloVe + SUP$^\dagger$ \citep{ethayarajh:2018:W18-30:uSIF} & \textbf{74.38} & \textbf{71.03} & \textbf{50.24} \\
\multicolumn{4}{l}{GloVe -- Considering Word Alignment} \\
\tblpad WMD$^\dagger$ \citep{pmlr-v37-kusnerb15:from} & 67.11 & 52.19 & 45.04 \\
\tblpad WMD$^\dagger$ {\small w/o stopwords} & 72.02 & 70.05 & 42.41 \\
\tblpad DynaMax \citep{zhelezniak2019:goforthemax:fuzzyjaccard} & 70.9 & - & - \\
\tblpad BERTScore$^\dagger$ \citep{zhang2019bertscore} & 67.26 & 50.93 & 44.77 \\
\tblpad \textbf{WRD} & \textsl{68.80} & \textsl{54.03} & \textsl{43.86} \\
\tblpad \textbf{WRD} + \textbf{VC}(WR) & \textsl{74.23} & \textsl{66.82} & \textsl{49.35} \\
\tblpad \textbf{WRD} + \textbf{VC}(SUP) & \textsl{77.03} & \textsl{72.66} & \textsl{55.90} \\
\tblpad \textbf{WRD} + \textbf{VC}(SWC) & \textsl{\uline{\textbf{77.63}}} & \textsl{\uline{\textbf{73.14}}} & \textsl{\uline{\textbf{56.81}}} \\
\midrule
\multicolumn{1}{l}{fastText -- Additive Composition} \\
\tblpad fastText$^\dagger$ & 67.85 & 60.95 & 51.42 \\
\tblpad fastText + WR$^\dagger$ \citep{arora2017simplebut:sif} & 72.15 & 69.48 & 48.76 \\
\tblpad fastText + SUP$^\dagger$ \citep{ethayarajh:2018:W18-30:uSIF} & \textbf{76.22} & \textbf{74.24} & \textbf{53.70} \\
\multicolumn{4}{l}{fastText -- Considering Word Alignment} \\
\tblpad WMD$^\dagger$ \citep{pmlr-v37-kusnerb15:from} & 67.58 & 52.31 & 44.34 \\
\tblpad WMD$^\dagger$ {\small w/o stopwords} & 71.61 & 69.41 & 40.94 \\
\tblpad DynaMax \citep{zhelezniak2019:goforthemax:fuzzyjaccard} & 76.6 & - & - \\
\tblpad BERTScore$^\dagger$ \citep{zhang2019bertscore} & 69.00 & 53.86 & 52.95 \\
\tblpad \textbf{WRD} & \textsl{73.31} & \textsl{62.10} & \textsl{56.70} \\
\tblpad \textbf{WRD} + \textbf{VC}(WR) & \textsl{76.81} & \textsl{71.94} & \textsl{54.93} \\
\tblpad \textbf{WRD} + \textbf{VC}(SUP) & \textsl{77.41} & \textsl{\uline{\textbf{76.97}}} & \textsl{57.54} \\
\tblpad \textbf{WRD} + \textbf{VC}(SWC) & \textsl{\uline{\textbf{77.85}}} & \textsl{74.94} & \textsl{\uline{\textbf{58.22}}} \\
\midrule
\tblpad Sent2Vec \citep{pagliardini2018sent2vec} & - & 75.5$^\ast$ & - \\
\tblpad Skip-Thought$^\ddagger$ \citep{kiros2015:skip-thought} & 46 & - & - \\
\tblpad ELMo $^\ddagger$ \citep{peters2018elmo} & 68 & - & - \\
\bottomrule
\end{tabular}
\egroup

\caption{%
    Pearson's $r \times 100$ between the predicted and gold scores is shown.
    The best results in each dataset, word vector, and strategy for computing the textual similarity (``Additive composition'' or ``Considering Word Alignment'') are in \textbf{bold};
    and the best results regardless of the strategy are further \uline{\textbf{underlined}}.
    Each row marked ($\dagger$) was re-implemented by us.
    Each value marked ($\ddagger$) was taken from \citet{perone2018:evaluationofsentenceembeddings},
    and marked ($\ast$) was taken from STS Wiki\protect\footnotemark.
}
\label{tab:benchmark}
\end{table}
\footnotetext{\url{http://ixa2.si.ehu.es/stswiki/index.php/STSbenchmark}}

Finally, we compared the performance of the proposed WRD and VC methods to that of various baselines, including recent alignment-based methods, i.e., \textbf{WMD} \cite{pmlr-v37-kusnerb15:from}, \textbf{BERTScore} \cite{zhang2019bertscore}\footnote{BERTScore was used as an STS method. We did not use BERT itself. See Appendix \ref{sec:bert_for_unsupervised_sts} for details.}, and \textbf{DynaMax} \cite{zhelezniak2019:goforthemax:fuzzyjaccard}.
The results are shown in Table~\ref{tab:benchmark}.
We summarize our major findings as follows.

\begin{itemize}
    \item 
    Among the methods that consider word alignment, WRD + VC achieved the best performance.
    This is likely due to the fact that other methods employ Euclidean distance (WMD) or dot product (DynaMax) as word similarity measures. These metrics cannot distinguish the two types of information (i.e., weight and dissimilarity).
    BERTScore applies cosine similarity like WRD; however, BERTScore was inferior to WRD on average, which can be attributed to the fact that BERTScore completely disregards the norm.

    \item 
    Compared to strong baselines based on additive composition (+WR, +SUP), WRD using the same word vectors (+\textbf{VC}(WR), +\textbf{VC}(WR)) performed equally or better.
    This result was unexpected given that +WR and +SUP were originally proposed to create sentence vectors, and WRD simply uses them without tuning.
    Thus, we believe that considering word alignment is an inherently good principle for STS.

    \MaybeCanOmit{%
    }
\end{itemize}

Refer to Appendix~\ref{sec:full_results} for the more comprehensive results obtained using additional datasets and methods, including (semi-)supervised approaches.

\section{Connection to Other Methods}
Finally, we discuss the relationships among \emph{WRD}, \emph{WMD}, and cosine similarity of additive composition (\ref{eq:additive_composition}, \ref{eq:additive_composition_cosine}), which we refer to as \emph{ADD}, from a sentence representation perspective.

\paragraph{Connection to Additive Composition.}
Surprisingly, ADD is a special case (i.e., a simplified version) of WRD.
In fact, given a discrete-distribution representation containing only a single sentence vector, i.e., $\VEC \mu_s^{\mathrm{point}} = \{(\VEC s, 1)\}$, the obvious EMD cost using cosine distance is equivalent to ADD.
\begin{align}
    \!\!\mathrm{EMD}(\VEC\mu_{s}^{\mathrm{point}}, \VEC\mu_{s'}^{\mathrm{point}};c^{}_{\cos}) = 1 - \cos(\VEC s, \VEC s')\text{.}\!\!
\end{align}
This relationship between ADD and WRD becomes clearer when examining their sentence representations using the norm ($\lambda_i$) and direction vector ($\VEC u_i$):%
\begin{subequations}\label{eq:sent_repr_wrd}
\begin{gather}
    \!\!\!\!
    \VEC s_{\textsc{add}}^{} \!\!\propto\!\! \frac 1 Z \! \sum_i \lambda^{}_i \VEC u^{}_i,
    \;
    \VEC \nu_{\textsc{WRD}}^{} \!\!=\!\! \frac 1 Z \!\sum_i \lambda^{}_i \Dirac{\VEC u^{}_i}
    \text{,}
    \!\!
    \tag{\theequation \ a,b}
\end{gather}
\end{subequations}
where $Z:=\sum_i\lambda_i$, and $\Dirac{\cdot}$ is the Dirac delta function.
Initially, they appear quite similar.
However, the key difference is that ADD treats a sentence as a single vector (the barycenter of direction vectors), whereas WRD treats a sentence as a set of direction vectors.
Given that STS tasks require word alignment (where words are treated disjointly), it is natural that WRD (where word vectors are treated disjointly) achieves better performance on STS tasks\footnote{%
In contrast, we have confirmed that ADD demonstrated higher performance than WRD on the topic similarity task (SICK-R. See Appendix~\ref{sec:full_results} for  details).
For a task where it is sufficient to know the trend of the meaning of the entire sentence, it may be preferable to aggregate the meaning of the entire sentence into a single vector.
}.

\paragraph{Connection to WMD.}
Why do WMD and WRD differ in performance on STS tasks even though both represent sentences as bag-of-word-vectors representations?
Sentence representations for ADD and WMD are as follows:
\begin{subequations}\label{eq:sent_repr_wmd}
\begin{gather}
    \!\!\!\!
    \VEC s_{\textsc{add}}^{} \!\!=\!\! \frac 1 n \!\sum_i 1 \!\cdot\! \VEC w^{}_i,
    \;
    \VEC \mu_{\textsc{WMD}}^{} \!\!=\!\! \frac 1 n \!\sum_i 1 \!\cdot\! \Dirac{\VEC w^{}_i}
    \text{.}
    \!\!\!
    \tag{\theequation \ a,b}
\end{gather}
\end{subequations}
The barycenters (\ref{eq:sent_repr_wrd}a), (\ref{eq:sent_repr_wmd}a) for ADD are identical up to scale because $\lambda_i^{} \VEC u_i^{} = \VEC w_i^{}$ holds.
In contrast, the discrete distributions (\ref{eq:sent_repr_wrd}b) for WRD and (\ref{eq:sent_repr_wmd}b) for WMD are quite different.
WRD treats the norm $\lambda$ as a weighting factor, as ADD implicitly does (\ref{eq:sent_repr_wmd}a). In contrast, WMD assigns uniform weights to both long and short vectors (\ref{eq:sent_repr_wmd}b), which is one reason the most natural representation (\ref{eq:sent_repr_wmd}b) employed by WMD does not work effectively.

The difference in performance between WMD and WRD can also be explained by the difference in the transportation cost functions.
Many word embeddings use inner product as the training objective function, i.e., the origin of the embedding space is meaningful.
Also, cosine distance used in WRD depends on the position of the origin.
In contrast, parallel translation invariant Euclidean cost used in WMD ignores the position of the origin.

\section{Conclusion}
In this paper, we first indicated (i) that the \emph{norm} and \emph{angle} of word vectors are good proxies for the importance of a word and dissimilarity between words, respectively, and (ii) that some previous alignment-based STS methods inappropriately ``mix up'' them.
With these findings, we have proposed word rotator's distance (WRD), which is a new unsupervised, EMD-based STS metric.
WRD was designed so that the norm and angle of word vectors correspond to the probability mass and transportation cost in EMD, respectively.
In addition, we found that the latest powerful sentence-vector estimation methods implicitly improve the norm and angle of word vectors, and we can exploit this effect as a word vector converter (VC).
In experiments on multiple STS tasks, the proposed methods outperformed not only alignment-based methods such as word mover's distance, but also powerful addition-based sentence vectors.

\newpage
\section*{Acknowledgments}
We appreciate the helpful comments from the anonymous reviewers.
We thank Emad Kebriaei for indicating how to normalize the sentence vectors.
We also thank
Benjamin Heinzerling, Masashi Yoshikawa, Hiroki Ouchi, Sosuke Kobayashi,
Paul Reisert, Ana Brassard, and Shun Kiyono
for constructive comments on the manuscript;
and Masatoshi Suzuki and Goro Kobayashi
for technical support.
This work was supported by JST CREST, Grant Number JPMJCR1513, Japan.
This work was also supported by JSPS KAKENHI, Grant Numbers JP19J21913 and JP19H04162.
\bibliography{Mendeley,mybib}
\bibliographystyle{acl_natbib}

\newpage
\appendix

\section{Resources Used in Experiments}
\label{sec:resources_experiments}
\subsection{Pre-trained Word Embeddings}
We used the following English pre-trained word embeddings in our experiments.
\begin{itemize}
    \item \textbf{GloVe} trained with Common Crawl~\citep{pennington-socher-manning:2014:EMNLP2014:glove}\footnote{\url{https://nlp.stanford.edu/projects/glove/}}
    \item \textbf{word2vec} trained with Google News~\citep{mikolov:2013:word2vec}\footnote{\url{https://code.google.com/archive/p/word2vec/}}
    \item \textbf{fastText} trained with Common Crawl~\citep{bojanowski2017fasttext}\footnote{\url{https://fasttext.cc/docs/en/english-vectors.html}}
    \item \textbf{PSL}, the ParagramSL-999 embeddings, trained with the PPDB paraphrase database~\citep{wieting2015frompara:psl}\footnote{\url{http://www.cs.cmu.edu/~jwieting/}}
    \item \textbf{ParaNMT} trained with ParaNMT-50, a large scale paraphrase database~\citep{wieting2018paranmt}\footnote{\url{https://github.com/kawine/usif}}
\end{itemize}

\subsection{Word Similarity Datasets}
We used the following nine word similarity tasks in our experiments.
\begin{itemize}
    \item \textbf{MEN}~\cite{bruni2012:men}
    \item \textbf{MTurk287}~\cite{radinsky2011:mturk287}
    \item \textbf{MC30}~\cite{millerandcharles1991:mc30}
    \item \textbf{RW}~\cite{luong2013:rw}
    \item \textbf{RG65}~\cite{rubensteinandgoodenough1965:rg65}
    \item \textbf{SCWS}~\cite{huang2012:scws}
    \item \textbf{SimLex999}~\cite{hill2015simlex999}
    \item \textbf{WS353}~\cite{finkelstein2001wordsim353}
\end{itemize}

\subsection{STS Datasets}
We used the following English STS datasets in our
experiments.

\begin{itemize}
    \item \textbf{STS'12}~\citep{agirre2012semeval}, \textbf{STS'13}~\citep{agirre2013starsem}, \textbf{STS'14}~\citep{agirre2014semeval}, and \textbf{STS'15}~\citep{agirre2015semeval}: semantic textual similarity shared tasks in SemEval
    \item \textbf{STS-B}: semantic textual similarity benchmark~\citep{cer2017semeval:stsb}, which is the collection from SemEval STS tasks 2012--2017~\citep{agirre2012semeval,agirre2013starsem,agirre2014semeval,agirre2015semeval,agirre2016semeval,cer2017semeval:stsb}
    \item \textbf{Twitter}: paraphrase and semantic similarity in twitter (PIT) task in SemEval 2015~\citep{xu2015semeval}
    \item \textbf{SICK-R}: SemEval 2014 semantic relatedness task~\citep{marelli2014semeval}
\end{itemize}

\paragraph{Tokenization.}
In each experiment, we first tokenized all the STS datasets (besides the Twitter dataset)
by NLTK~\citep{bird-loper-2004-nltk} with some post-processing steps following \citet{ethayarajh:2018:W18-30:uSIF}\footnote{\url{https://github.com/kawine/usif}}.
The Twitter dataset has already been tokenized by the workshop organizer.
We then lowercased all tokens to conduct experiments under the same conditions with cased embeddings and non-cased embeddings.

\subsection{Stopword List}
The stopword list based on the SMART Information Retrieval System\footnote{\url{https://github.com/igorbrigadir/stopwords}} was used for WMD~\cite{pmlr-v37-kusnerb15:from} and conceptor removal (C)~\cite{liu2019NAACLconceptor}.

\section{%
Contextualized Word Embeddings on Unsupervised STS
}
\label{sec:bert_for_unsupervised_sts}

BERT~\cite{devlin2018bert} and its variants have not yet shown good results on \emph{unsupervised} STS (note that, in a supervised or semi-supervised setting where there exists training data or external resources, BERT-based models show the current, best results).
One particularly promising usage of BERT-based models for unsupervised STS is BERTScore~\cite{zhang2019bertscore}, which was originally proposed as an automatic evaluation metric.
However, our preliminary experiments\footnote{%
    We used BERT-large and RoBERTa-large.
    For embeddings, we used either the last layer or the concatenation of all the layers.
    In the original paper, which allows the use of teacher data, the development set was used to select the layer.
} show that BERTScore with pre-trained BERT/RoBERTa performs poorly on unsupervised STS.

Nonetheless, BERTScore is definitely promising as a method. We then reported the results of BERTScore using \emph{non-}contextualized word vectors, e.g., GloVe, and we confirmed higher performance compared to using pre-trained BERT.
Needless to say, the application of BERT-based models to unsupervised STS is an important future research topic.

\newpage
\section{Algorithm of Word Rotator's Distance}
\label{sec:appendix:algorithm_of_wrd}
The algorithm used in the actual computation of WRD is shown in Algorithm~\ref{alg:WRD}.
\begin{algorithm}[t]
\caption{Word Rotator's Distance (WRD)}
\label{alg:WRD}
\begin{algorithmic}[1]
\Input{%
    a pair of sentences $s=(\VEC w_1,\dots,\VEC w_n)$, $s'=(\VEC w'_1,\dots,\VEC w'_{n'})$
    }
    \Let{Z}{\sum_{i=1}^n \norm{\VEC w^{}_i} \in \mathbb R}
    \Let{Z'}{\sum_{j=1}^{n'} \norm{\VEC w'_j} \in \mathbb R}
    \Let{\VEC m^{}_{s}}{\frac 1 Z(\norm{\VEC w^{}_1},\dots,\norm{\VEC w^{}_n}) \in \mathbb R^n}
    \Let{\VEC m^{}_{s'}}{\frac 1 {Z'}(\norm{\VEC w'_1},\dots,\norm{\VEC w'_{n'}}) \in \mathbb R^{n'}}
    \For{$i \gets 1$ to $n$}
        \For{$j \gets 1$ to $n'$}
            \Let{\VEC C^{}_{ij}}{1 - \cos(\VEC w_i^{}, \VEC w'_j)}
        \EndFor
    \EndFor
\Let{\textsc{WRD}(s,s')}{\textsc{EMD}(\VEC m^{}_{s}, \VEC m^{}_{s'}; \VEC C)}
\Output{
$\textsc{WRD}(s,s')\in \mathbb R$
}
\end{algorithmic}

\end{algorithm}

For EMD computation, off-the-shelf libraries can be used\footnote{%
In our experiments, we used the well-developed python optimal transport (POT) library~\cite{flamary2017pot}: \url{https://github.com/rflamary/POT/}. In particular, \texttt{ot.emd2()} was used.%
}.
Note that most EMD (optimal transport) libraries take two probabilities (mass) $\VEC m \in \mathbb R^n,\, \VEC m'\in \mathbb R^{n'}$ and a cost matrix $\VEC C \in \mathbb R^{n\times n'}$ with $\VEC C^{}_{ij} = d(\VEC x^{}_i, \VEC x'_j)$ as inputs. Parameters ($\VEC m$, $\VEC m'$, $\VEC C$) have the same information as ($\VEC \mu$, $\VEC \mu'$, $d$), introduced in Section~\ref{sec:problem_of_wmd}.
The notation of Algorithm \ref{alg:WRD} follows this style.

The cosine distance $1 - \cos(\VEC w^{}_i,\VEC w'_j)$ in line 7 of Algorithm~\ref{alg:WRD} is equivalent to $1 - \cos(\VEC u^{}_i, \VEC u'_j)$ in Equation~\ref{eq:cosine_distance}.
We adopted the former simply to reduce the computation steps.

\section{Algorithms of Vector Converter}
\label{sec:appendix:algorithm_of_vc}
\begin{algorithm}[!h]
\caption{Vector Converter (VC), induced from 
\textbf{A}ll-but-the-top \citep{mu2018allbutthetop},
SIF \textbf{W}eighting \citep{arora2017simplebut:sif}
or \textbf{U}nsupervised SIF \citep{ethayarajh:2018:W18-30:uSIF},
and common component \textbf{R}emoval \citep{arora2017simplebut:sif}
or \textbf{P}iecewise CCR \citep{ethayarajh:2018:W18-30:uSIF}
or \textbf{C}onceptor removal \citep{liu2019NAACLconceptor}
.}
\label{alg:VC}
\begin{algorithmic}[1]
\Input{%
pre-trained word vectors $\{\VEC w_1^{},\dots,\VEC w_{\abs{\mathcal V}}\}$,
sentences in interest $\mathcal S = \{s_1^{},\dots,s_{\abs{\mathcal S}}^{}\}$,
word probability $\mathbb P\colon$ $\mathcal V\!\to\!\mathcal [0,1]$,
stopwords $\mathcal V^{}_{\mathrm{SW}}\subseteq\mathcal V$,
and constants
$D^{}_{\mathrm{A}}$,
$a^{}_{\mathrm{W}}$ (or $a^{}_{\mathrm{U}}$),
$D^{}_3$ (or $D^{}_{\mathrm{C}}$)
}
\Statex{\uline{Compute parameters of $f^{}_1$:}}
\Statex{$\cdots$ if using \textbf{A}ll-but-the-top:}
    \Let{\overline{\VEC w}}{\frac{1}{\abs{\mathcal V}}\sum_{i=1}^{\abs{\mathcal V}} \VEC w_i \in \mathbb R^d}
    \For{$i \gets 1$ to $\abs{\mathcal V}$}
    \Let{\overline{\VEC w}_i}{\VEC w_i - \overline{\VEC w}}
    \EndFor
    \Let{\VEC u^{}_1,\dots,\VEC u^{}_{D^{}_{\mathrm{A}}}}{\mathrm{PCA}(\{\overline{\VEC w}_1^{},\dots,\overline{\VEC w}_{\abs{\mathcal V}}^{}\})}
    \Statex{\Comment{top $D^{}_{\mathrm{A}}$ singular vectors}}
    \Let{\VEC A^{}_1}{\VEC I - \sum_{j=1}^{D^{}_{\mathrm{A}}} \VEC u^{}_j \VEC u_j^\top\in \mathbb R^{d\times d}}
    \Let{\VEC b^{}_1}{\overline{\VEC w}\in\mathbb R^d}
\Statex{\uline{Compute parameters of $\alpha^{}_2$:}}
\ForEach {$w$}
\Statex{\quad\; $\cdots$ if using SIF \textbf{W}eighting:}
\Let{\alpha^{}_2(w)}{a^{}_{\mathrm{W}} / (\mathbb P(w) + a^{}_{\mathrm{W}}) \in \mathbb R}
\Statex{\quad\; $\cdots$ else if using \textbf{U}nsupervised SIF:}
\Let{\alpha^{}_2(w)}{a^{}_{\mathrm{U}} / (\mathbb P(w) + \frac 1 2 a^{}_{\mathrm{U}}) \in \mathbb R}
\EndFor
\Statex{\uline{Create sentence vectors temporarily:}}
    \For{$i \gets 1$ to $\abs{\mathcal S}$}
        \Let{\VEC s_i^{}\!}{\!\frac 1 {\abs{s^{}_i}}\sum_{w\in s_i^{}}\alpha_2^{}(w)\VEC A_1(\VEC w - \VEC b^{}_1) \in \mathbb R^d}
    \EndFor
\Statex{\uline{Compute parameters of $f^{}_3$:}}
\Statex{$\cdots$ if using CC\textbf{R} or \textbf{P}iecewise CCR:}
    \Let{\!(\VEC v^{}_1,\sigma^{}_1),\dots,\!(\VEC v^{}_{\!D^{}_3},\sigma^{}_{\!D^{}_3\!})\!}{\!\mathrm{PCA}(\{{\VEC s}_{}^{},\dots,\!{\VEC s}_{\abs{\mathcal S}}^{}\!\})\!\!\!\!\!}
        \Statex{\Comment{top $D^{}_3$ singular vectors and singular values}}
    \Let{\VEC A_3^{}}{\displaystyle \VEC I - \sum_{i=1}^{D^{}_3} \frac{\sigma_i^2}{\sum_{j=1}^{D^{}_3} \sigma_j^2} \,\VEC v^{}_{i} \VEC v_{i}^\top \in \mathbb R^{d\times d}}
\Statex{$\cdots$ else if using \textbf{C}onceptor removal:}
    \Let{\VEC R}{\frac 1 {\abs{\mathcal S}} \sum_{i=1}^{\abs{\mathcal S}} \VEC s^{}_i \VEC s_i^\top \in \mathbb R^{d\times d}}
    \Let{\VEC C}{\VEC R(\VEC R + \alpha_{\mathrm{C}}^{-2} \VEC I)^{-1} \in \mathbb R^{d\times d}}
    \Let{\VEC R^{}_{\mathrm{SW}}}{\frac 1 {\abs{\mathcal V^{}_{\mathrm{SW}}}} \sum_{w\in\mathcal V^{}_{\mathrm{SW}}} \VEC w_{}^{} \VEC w_{}^\top}
    \Let{\VEC C^{}_{\mathrm{SW}}}{\VEC R^{}_{\mathrm{SW}}(\VEC R^{}_{\mathrm{SW}} + \alpha_{\mathrm{C}}^{-2} \VEC I)^{-1}}
    \Let{\VEC A^{}_3}{((\VEC I - \VEC C)^{-1} + (\VEC I - \VEC C^{}_{\mathrm{SW}})^{-1} - \VEC I)^{-1}}
\Statex{\uline{Convert word vectors:}}
\For{$i \gets 1$ to $\abs{\mathcal V}$}
    \Let{\widetilde{\VEC w}^{}_{i}}{\VEC A^{}_3( \alpha^{}_2(w) \VEC A^{}_1(\VEC w^{}_i - \VEC b^{}_1))}
\EndFor
\Output{Converted word vectors $\{\widetilde{\VEC w}^{}_1,\cdots,\widetilde{\VEC w}^{}_{\abs{\mathcal V}}\}$}
\end{algorithmic}

\end{algorithm}
Algorithm \ref{alg:VC} summarizes the overall procedure of word vector converter $f^{}_{\mathrm{VC}}$~(Equation \ref{eq:word_vector_converter}).

When computing Algorithm \ref{alg:VC},
we set hyperparameters as
\begin{itemize}
    \item $D^{}_{\mathrm{A}} = 3$ for \textbf{A} \cite{mu2018allbutthetop}
    \item $a^{}_{\mathrm{W}} = 10^{-3}$ for \textbf{W} \cite{arora2017simplebut:sif}
    \item $a^{}_{\mathrm{U}} \approx 1.2 \times 10^{-3}$ for \textbf{U} \cite{ethayarajh:2018:W18-30:uSIF}
    \item $D^{}_3 = 1$ for \textbf{R} \cite{arora2017simplebut:sif}
    \item $D^{}_3 = 5$ for \textbf{P} \cite{ethayarajh:2018:W18-30:uSIF}
    \item $D^{}_{\mathrm C} = 1$ for \textbf{C} \cite{liu2019NAACLconceptor}
\end{itemize}
reported in previous studies, without any additional tuning.

Prior to performing all-but-the-top (A), we restricted the vocabulary of word vectors to words appearing more than 200 times in the enwiki corpus\footnote{\url{https://github.com/PrincetonML/SIF/}. Preprocessed by \citet{arora2017simplebut:sif}.}, following \citet{liu2019AAAIconceptor}.

We used the unigram probability $\mathbb P$ of English words estimated using the enwiki dataset, preprocessed by \citet{arora2017simplebut:sif}\footnote{\url{https://github.com/PrincetonML/SIF/}}.

See Table~\ref{tb:affine_sentence_encoders} for an overview of the existing methods.
\label{sec:appendix_unsupervised_sentence_encoders}
\begin{table*}
\begin{center}
\setlength{\tabcolsep}{1pt}  %
\footnotesize
\begin{tabular}{llll}
\toprule
 & $f_1^{}$ & $\alpha_2^{}$ & $f_3^{}$ \\
 & denoising word vectors & scaling & denoisng sentence vectors \\
\midrule
well-known heuristic & --  & Stop Words Removal & -- \\
well-known heuristic & --  & IDF {\scriptsize (Inverse Document Frequency)} & -- \\
\midrule[0.3pt]
\citet{arora2017simplebut:sif} & -- & SIF {\scriptsize (Smoothed Inverse Frequency)} & Common Component Removal \\
\citet{mu2018allbutthetop} & all-but-the-top & -- & -- \\
\citet{ethayarajh:2018:W18-30:uSIF} & {Sentence-wise Feature Scaling} & uSIF {\scriptsize (Unsupervised SIF)} & Piecewise CCR \\
\citet{liu2019AAAIconceptor} & Conceptor Negation & -- & -- \\
\citet{liu2019NAACLconceptor} & -- & SIF & Conceptor Removal \\
\bottomrule
\end{tabular}

\caption{%
Sentence-vector estimation methods.
}
\label{tb:affine_sentence_encoders}
\end{center}
\end{table*}%
There are many possible combinations of $f_1$, $f_2$, and $f_3$, and exploring them is a good direction for future work.

\section{Full Results of Experiments}
\label{sec:full_results}

\subsection{Workings of Angle}
See Table~\ref{tb:workings_of_angle_full} for full results using nine word similarity datasets.
\label{sec:full_results_angle}
\begin{table}[tb]
\centering
\small
\setlength{\tabcolsep}{2.0pt}  %
\renewcommand{\arraystretch}{1.15} %
\newcommand{\tblpaddummy}{\hspace{3pt}}
\begin{subtable}[t]{.54\linewidth}
    \centering
    \begin{tabular}[t]{lccc}
    \toprule
     & \textsc{L2} & \textsc{dot} & \textsc{cos} \\
    \vphantom{\makecell{+ \textbf{VC} \\ (AWR)}} &&& \\
    \midrule
    GloVe &&& \\
    \tblpaddummy MEN       & 73.36 & \textbf{80.79} & \textbf{80.49} \\
    \tblpaddummy MTurk287  & 60.87 & \textbf{69.50} & \textbf{69.18} \\
    \tblpaddummy MC30      & 75.22 & 76.77          & \textbf{78.81} \\
    \tblpaddummy RW        & 45.13 & 48.17          & \textbf{51.04} \\
    \tblpaddummy RG65      & 70.75 & \textbf{77.79} & 76.90          \\
    \tblpaddummy SCWS      & 56.25 & 62.20          & \textbf{63.29} \\
    \tblpaddummy SimLex999 & 35.03 & 38.86          & \textbf{40.71} \\
    \tblpaddummy WS353-REL & 49.74 & \textbf{72.35} & 68.75          \\
    \tblpaddummy WS353-SIM & 69.03 & \textbf{79.54} & \textbf{79.57} \\
    \midrule
    word2vec &&& \\
    \tblpaddummy MEN       & 62.31 & 74.46 & \textbf{78.20} \\
    \tblpaddummy MTurk287  & 49.43 & 66.66 & \textbf{68.37} \\
    \tblpaddummy MC30      & 69.88 & 76.57 & \textbf{78.87} \\
    \tblpaddummy RW        & 35.10 & 51.29 & \textbf{55.80} \\
    \tblpaddummy RG65      & 71.30 & 72.58 & \textbf{76.17} \\
    \tblpaddummy SCWS      & 47.46 & 65.10 & \textbf{66.49} \\
    \tblpaddummy SimLex999 & 32.35 & 43.23 & \textbf{44.20} \\
    \tblpaddummy WS353-REL & 40.74 & 56.65 & \textbf{61.40} \\
    \tblpaddummy WS353-SIM & 55.82 & 74.89 & \textbf{77.39} \\
    \midrule
    fastText &&& \\
    \tblpaddummy MEN       & 68.67 & 79.37 & \textbf{84.55} \\
    \tblpaddummy MTurk287  & 56.74 & 66.10 & \textbf{72.47} \\
    \tblpaddummy MC30      & 77.11 & 81.36 & \textbf{85.24} \\
    \tblpaddummy RW        & 47.24 & 53.83 & \textbf{62.17} \\
    \tblpaddummy RG65      & 79.72 & 80.58 & \textbf{86.26} \\
    \tblpaddummy SCWS      & 52.88 & 64.90 & \textbf{69.42} \\
    \tblpaddummy SimLex999 & 36.91 & 46.71 & \textbf{50.47} \\
    \tblpaddummy WS353-REL & 49.44 & 63.64 & \textbf{72.33} \\
    \tblpaddummy WS353-SIM & 66.83 & 79.34 & \textbf{84.24} \\
    \bottomrule
    \end{tabular}
    \caption{It is better to use only direction instead of the norm.}
    \label{subtb:workings_of_angle_full_cos}
\end{subtable}
\hspace{0.03\linewidth}
\begin{subtable}[t]{.40\linewidth}
    \centering
    \begin{tabular}[t]{ccc}
    \toprule
    \textsc{cos} & \textsc{cos} & \textsc{cos} \\
    \makecell{original \\ vector} & + A & \makecell{+ \textbf{VC} \\ (AWR)} \\
    \midrule
    \multicolumn{3}{l}{GloVe} \\
    \tblpaddummy 80.49          & 82.58 & \textbf{83.69} \\
    \tblpaddummy 69.18          & \textbf{73.11} & 72.73 \\
    \tblpaddummy \textbf{78.81} & 77.29 & \textbf{78.51} \\
    \tblpaddummy 51.04          & \textbf{57.87} & \textbf{57.96} \\
    \tblpaddummy \textbf{76.90} & 76.30 & 76.28 \\
    \tblpaddummy 63.29          & \textbf{66.65} & \textbf{66.48} \\
    \tblpaddummy 40.71          & \textbf{46.55} & \textbf{47.02} \\
    \tblpaddummy 68.75          & \textbf{72.27} & \textbf{72.14} \\
    \tblpaddummy 79.57          & \textbf{81.19} & 80.29 \\
    \midrule
    \multicolumn{3}{l}{word2vec} \\
    \tblpaddummy 78.20 & \textbf{80.22} & \textbf{80.15} \\
    \tblpaddummy 68.37 & \textbf{65.21} & \textbf{65.20} \\
    \tblpaddummy 78.87 & 81.06 & \textbf{82.20} \\
    \tblpaddummy 55.80 & \textbf{57.39} & \textbf{57.47} \\
    \tblpaddummy 76.17 & \textbf{82.22} & \textbf{82.57} \\
    \tblpaddummy 66.49 & \textbf{65.08} & \textbf{65.23} \\
    \tblpaddummy 44.20 & \textbf{46.23} & \textbf{46.53} \\
    \tblpaddummy 61.40 & \textbf{60.96} & \textbf{61.07} \\
    \tblpaddummy 77.39 & \textbf{77.07} & \textbf{77.18} \\
    \midrule
    \multicolumn{3}{l}{fastText} \\
    \tblpaddummy 84.55 & \textbf{85.51} & \textbf{85.91} \\
    \tblpaddummy \textbf{72.47} & \textbf{71.25} & 71.02 \\
    \tblpaddummy 85.24 & \textbf{86.74} & \textbf{86.69} \\
    \tblpaddummy 62.17 & \textbf{62.98} & \textbf{62.75} \\
    \tblpaddummy 86.26 & 88.58 & \textbf{89.33} \\
    \tblpaddummy \textbf{69.42} & \textbf{69.07} & 68.62 \\
    \tblpaddummy 50.47 & \textbf{51.83} & \textbf{52.26} \\
    \tblpaddummy \textbf{72.33} & \textbf{72.32} & \textbf{72.81} \\
    \tblpaddummy \textbf{84.24} & \textbf{84.69} & \textbf{84.46} \\
    \bottomrule
    \end{tabular}
    \caption{%
    VC gradually ``grows'' the direction of word vectors.}
    \label{subtb:workings_of_angle_full_vc}
\end{subtable}
\caption{%
    The angle of word vectors is a good proxy for word similarity (full results).
    Spearman's $\rho$ $\times$ $100$
    between the predicted and gold scores is reported.
    In each row, the best result and results where the difference from the best result was $<0.5$ are indicated in \textbf{bold}.
    ``+ AW'' is omitted from Table~\ref{subtb:workings_of_angle_vc} because \textsc{W} (i.e., the scaling function) alone does not change the angle.
    }
\label{tb:workings_of_angle_full}
\end{table}
\section{Full Results of Ablation Study}
\label{sec:ablation_results}
See Table~\ref{tb:ablation_full} for full results.
WRD \emph{without} stopword removal achieves the best results.
This is likely because WRD can more continuatively compare the differences in the importance between stopwords using their norm.

\begin{table}
\centering
\setlength{\tabcolsep}{2pt}  %
\renewcommand{\arraystretch}{1.15} %
\small
\begin{tabular}{lccccc}
\addlinespace[-\aboverulesep] 
\cmidrule[\heavyrulewidth]{1-3}
\cmidrule[\heavyrulewidth]{5-6}
 & WMD & \textbf{WRD} & & WMD & \textbf{WRD} \\
Removing Stopwords & & & & \checkmark & \checkmark \\
\cmidrule{1-3}
\cmidrule{5-6}
GloVe & 65.74 & \textbf{67.70} && \textbf{75.42} & 74.43 \\
GloVe + A & 65.44 & \textbf{68.26} && \textbf{74.78} & \textbf{74.68} \\
GloVe + AW & 64.76 & \textbf{76.08} && 74.53 & \textbf{75.62} \\
GloVe + \textbf{VC}(AWR) & 64.74 & \uline{\textbf{76.87}} && 74.47 & \textbf{76.49} \\
GloVe + SIF weights & 75.42 & - && 76.41 & - \\
GloVe + A + SIF weights & 75.59 & - && 75.75 & - \\
\cmidrule{1-3}
\cmidrule{5-6}
word2vec & 67.21 & \textbf{70.91} && 72.40 & \textbf{73.11} \\
word2vec + A & 67.09 & \textbf{71.23} && 72.31 & \textbf{73.46} \\
word2vec + AW & 64.97 & \textbf{75.00} && 71.88 & \textbf{74.25} \\
word2vec + \textbf{VC}(AWR) & 64.89 & \uline{\textbf{76.04}} && 71.71 & \textbf{75.39} \\
word2vec + SIF weights & 73.90 & - && 73.41 & - \\
word2vec + A + SIF weights & 73.76 & - && 73.27 & - \\
\cmidrule{1-3}
\cmidrule{5-6}
fastText & 64.06 & \textbf{69.31} && 73.82 & \textbf{75.65} \\
fastText + A & 63.79 & \textbf{69.34} && 73.32 & \textbf{75.72} \\
fastText + AW & 62.47 & \textbf{76.90} && 72.89 & \textbf{76.54} \\
fastText + \textbf{VC}(AWR) & 62.47 & \uline{\textbf{77.56}} && 72.85 & \textbf{77.26} \\
fastText + SIF weights & 74.64 & - && 74.79 & - \\
fastText + A + SIF weights & 74.28 & - && 74.28 & - \\
\cmidrule[\heavyrulewidth]{1-3}
\cmidrule[\heavyrulewidth]{5-6}
\addlinespace[-\belowrulesep]
\end{tabular}

\caption{%
    The combination of WRD and VC gave the best performance (full results). Pearson's $r \times 100$ between the predicted and gold scores is reported.
    The STS-B dataset (dev) was used.
    The best result and results where the difference from the best $<0.5$ in each row are in \textbf{bold},
    and the best result in each word vector is further \uline{\textbf{underlined}}.
}
\label{tb:ablation_full}
\end{table}

\section{Full Results of Comparative Experiments}
\label{sec:full_results_benchmark}
See Table~\ref{tab:benchmark2} for full results in an unsupervised settings.
See Table~\ref{tab:benchmark3} for full results in an semisupervised and supervised settings.

\begin{table*}
\centering
\bgroup
\setlength{\tabcolsep}{2pt}
\renewcommand{\arraystretch}{0.94}
\footnotesize
\begin{tabular}{@{}l ccccc ccccc@{}}
\toprule
& STS'12 & STS'13 & STS'14 & STS'15 & STS-B & Twitter & SICK-R \\
\midrule
\multicolumn{1}{l}{GloVe -- Additive Composition} \\
\tblpad GloVe$^\dagger$ & 53.04 & 45.52 & 57.97 & 56.08 & 45.57 & 29.35 & 66.79 \\
\tblpad GloVe + WR \citep{arora2017simplebut:sif} & 56.2 & 56.6 & 68.5 & 71.7 & - & 48.0 & 72.2 \\
\tblpad GloVe + WR$^\dagger$ \citep{arora2017simplebut:sif} & 60.57 & 54.99 & 67.74 & 67.74 & 62.85 & 40.03 & 69.32 \\
\tblpad GloVe + SUP \citep{ethayarajh:2018:W18-30:uSIF} & \uline{\textbf{64.9}} & \uline{\textbf{63.6}} & \uline{\textbf{74.4}} & \textbf{76.1} & \textbf{71.5} & - & \uline{\textbf{73.0}} \\
\tblpad GloVe + SUP$^\dagger$ \citep{ethayarajh:2018:W18-30:uSIF} & 64.85 & 62.50 & 73.69 & 74.38 & 71.03 & \textbf{50.24} & 72.34 \\
\multicolumn{1}{l}{GloVe -- Considering Word Alignment} \\
\tblpad WMD\; GloVe$^\dagger$ \citep{pmlr-v37-kusnerb15:from} & 55.74 & 44.18 & 60.24 & 67.11 & 52.19 & 45.04 & 61.91 \\
\tblpad WMD\; GloVe \small{w/o stopwords}$^\dagger$ \citep{pmlr-v37-kusnerb15:from} & 60.67 & 53.45 & 67.63 & 72.02 & 70.05 & 42.41 & 63.31 \\
\tblpad DynaMax\; GloVe \citep{zhelezniak2019:goforthemax:fuzzyjaccard} & 58.2 & 53.9 & 65.1 & 70.9 & - & - & - \\
\tblpad BERTScore\; GloVe$^\dagger$ \citep{zhang2019bertscore} & 52.81 & 47.23 & 62.06 & 67.26 & 50.93 & 44.77 & 65.28 \\
\tblpad \textbf{WRD}\; GloVe & \textsl{58.28} & \textsl{48.79} & \textsl{62.31} & \textsl{68.80} & \textsl{54.03} & \textsl{43.86} & \textsl{63.84} \\
\tblpad \textbf{WRD}\; GloVe + \textbf{VC}(WR) & \textsl{62.96} & \textsl{56.88} & \textsl{68.73} & \textsl{74.23} & \textsl{66.82} & \textsl{49.35} & \textsl{66.94} \\
\tblpad \textbf{WRD}\; GloVe + \textbf{VC}(SUP) & \textsl{64.28} & \textsl{58.19} & \textsl{71.10} & \textsl{77.03} & \textsl{72.66} & \textsl{55.90} & \textsl{67.29} \\
\tblpad \textbf{WRD}\; GloVe + \textbf{VC}(SWC) & \textsl{64.22} & \textsl{58.07} & \textsl{71.58} & \textsl{77.63} & \textsl{73.14} & \textsl{56.81} & \textsl{\textbf{67.79}} \\
\tblpad \textbf{WRD}\; GloVe + \textbf{VC}(SUC) & \textsl{\textbf{64.39}} & \textsl{\textbf{58.41}} & \textsl{\textbf{72.00}} & \textsl{\uline{\textbf{77.76}}} & \textsl{\uline{\textbf{74.16}}} & \textsl{\uline{\textbf{57.11}}} & \textsl{67.07} \\
\midrule
\multicolumn{1}{l}{word2vec -- Additive Composition} \\
\tblpad word2vec$^\dagger$ & 61.67 & 53.07 & 67.63 & 67.45 & 61.54 & 30.54 & \uline{\textbf{72.51}} \\
\tblpad word2vec + WR$^\dagger$ \citep{arora2017simplebut:sif} & 62.79 & \textbf{58.55} & 71.11 & 70.41 & 67.49 & \textbf{35.59} & 70.78 \\
\tblpad word2vec + SUP$^\dagger$ \citep{ethayarajh:2018:W18-30:uSIF} & \uline{\textbf{63.27}} & 58.50 & \uline{\textbf{71.72}} & \textbf{72.97} & \textbf{69.39} & 34.72 & 70.51 \\
\multicolumn{1}{l}{word2vec -- Considering Word Alignment} \\
\tblpad WMD\; word2vec$^\dagger$ \citep{pmlr-v37-kusnerb15:from} & 55.89 & 44.52 & 60.24 & 66.46 & 56.10 & 39.53 & 64.05 \\
\tblpad WMD\; word2vec \small{w/o stopwords}$^\dagger$ \citep{pmlr-v37-kusnerb15:from} & 58.14 & 49.95 & 65.22 & 70.54 & 67.46 & 36.00 & 62.41 \\
\tblpad DynaMax\; word2vec \citep{zhelezniak2019:goforthemax:fuzzyjaccard} & 53.7 & \uline{\textbf{59.5}} & 68.0 & 74.2 & - & - & - \\
\tblpad BERTScore\; word2vec$^\dagger$ \citep{zhang2019bertscore} & 47.83 & 43.54 & 56.26 & 62.06 & 49.16 & 34.07 & 58.75 \\
\tblpad \textbf{WRD}\; word2vec & \textsl{59.14} & \textsl{51.41} & \textsl{65.36} & \textsl{72.39} & \textsl{\uline{\textbf{72.39}}} & \textsl{41.44} & \textsl{66.31} \\
\tblpad \textbf{WRD}\; word2vec + \textbf{VC}(WR) & \textsl{61.45} & \textsl{55.98} & \textsl{68.52} & \textsl{74.86} & \textsl{70.13} & \textsl{43.42} & \textsl{\textbf{66.76}} \\
\tblpad \textbf{WRD}\; word2vec + \textbf{VC}(SUP) & \textsl{61.85} & \textsl{55.38} & \textsl{68.96} & \textsl{75.30} & \textsl{71.19} & \textsl{42.86} & \textsl{66.11} \\
\tblpad \textbf{WRD}\; word2vec + \textbf{VC}(SWC) & \textsl{\textbf{61.95}} & \textsl{55.47} & \textsl{69.18} & \textsl{\uline{\textbf{75.69}}} & \textsl{71.59} & \textsl{43.99} & \textsl{66.53} \\
\tblpad \textbf{WRD}\; word2vec + \textbf{VC}(SUC) & \textsl{61.95} & \textsl{55.64} & \textsl{\textbf{69.36}} & \textsl{75.56} & \textsl{72.29} & \textsl{\uline{\textbf{44.17}}} & \textsl{65.54} \\
\midrule
\multicolumn{1}{l}{fastText -- Additive Composition} \\
\tblpad fastText$^\dagger$ & 59.76 & 52.79 & 67.42 & 67.85 & 60.95 & 51.42 & 70.44 \\
\tblpad fastText + WR$^\dagger$ \citep{arora2017simplebut:sif} & 64.03 & 59.90 & 72.88 & 72.15 & 69.48 & 48.76 & \uline{\textbf{72.19}} \\
\tblpad fastText + SUP$^\dagger$ \citep{ethayarajh:2018:W18-30:uSIF} & \uline{\textbf{64.39}} & \uline{\textbf{62.33}} & \uline{\textbf{74.82}} & \textbf{76.22} & \textbf{74.24} & \textbf{53.70} & 72.13 \\
\multicolumn{1}{l}{fastText -- Considering Word Alignment} \\
\tblpad WMD\; fastText$^\dagger$ \citep{pmlr-v37-kusnerb15:from} & 55.27 & 44.39 & 60.09 & 67.58 & 52.31 & 44.34 & 62.21 \\
\tblpad WMD\; fastText \small{w/o stopwords}$^\dagger$ \citep{pmlr-v37-kusnerb15:from} & 60.00 & 52.29 & 66.87 & 71.61 & 69.41 & 40.94 & 62.84 \\
\tblpad DynaMax\; fastText \citep{zhelezniak2019:goforthemax:fuzzyjaccard} & 60.9 & \textbf{60.3} & 69.5 & 76.6 & - & - & - \\
\tblpad BERTScore\; fastText$^\dagger$ \citep{zhang2019bertscore} & 51.95 & 45.86 & 61.66 & 69.00 & 53.86 & 52.95 & 64.69 \\
\tblpad \textbf{WRD}\; fastText & \textsl{58.84} & \textsl{50.74} & \textsl{64.60} & \textsl{73.31} & \textsl{62.10} & \textsl{56.70} & \textsl{64.90} \\
\tblpad \textbf{WRD}\; fastText + \textbf{VC}(WR) & \textsl{63.50} & \textsl{58.44} & \textsl{70.26} & \textsl{76.81} & \textsl{71.94} & \textsl{54.93} & \textsl{\textbf{67.85}} \\
\tblpad \textbf{WRD}\; fastText + \textbf{VC}(SUP) & \textsl{\textbf{64.22}} & \textsl{58.84} & \textsl{71.41} & \textsl{77.41} & \textsl{\uline{\textbf{76.97}}} & \textsl{57.54} & \textsl{67.36} \\
\tblpad \textbf{WRD}\; fastText + \textbf{VC}(SWC) & \textsl{64.07} & \textsl{58.75} & \textsl{71.59} & \textsl{\uline{\textbf{77.85}}} & \textsl{74.94} & \textsl{\uline{\textbf{58.22}}} & \textsl{67.83} \\
\tblpad \textbf{WRD}\; fastText + \textbf{VC}(SUC) & \textsl{64.00} & \textsl{59.06} & \textsl{\textbf{71.81}} & \textsl{77.77} & \textsl{75.56} & \textsl{57.98} & \textsl{67.01} \\
\midrule
\tblpad Sent2Vec \citep{pagliardini2018sent2vec} & - & - & - & - & 75.5$^\ast$ & - & - \\
\tblpad Skip-Thought$^\ddagger$ \citep{kiros2015:skip-thought} & 41 & 29 & 40 & 46 & - & - & - \\
\tblpad ELMo (All layers, 5.5B)$^\ddagger$ \citep{peters2018elmo} & 55 & 53 & 63 & 68 & - & - & - \\
\bottomrule
\end{tabular}
\egroup

\caption{%
    Pearson's $r \times 100$ between the predicted scores and the gold scores is shown.
    The best results in each block
    is in \textbf{bold},
    and the best results regardless of the strategy for computing textual similarity are further \uline{\textbf{underlined}}.
    The results of our methods are \textsl{slanted}.
    Each row marked ($\dagger$) was re-implemented by us.
    Each value marked ($\ddagger$) was taken from \citet{perone2018:evaluationofsentenceembeddings}.
    Each value marked ($\ast$) was taken from STS Wiki (\url{http://ixa2.si.ehu.es/stswiki/index.php/STSbenchmark}).
}
\label{tab:benchmark2}
\end{table*}
\begin{table*}
\centering
\bgroup
\setlength{\tabcolsep}{2pt}
\renewcommand{\arraystretch}{0.94}
\footnotesize
\begin{tabular}{@{}l ccccc ccccc@{}}
\toprule
& STS'12 & STS'13 & STS'14 & STS'15 & STS-B & Twitter & SICK-R \\
\midrule
\multicolumn{1}{l}{\textbf{Semi-supervised}} \\
\multicolumn{1}{l}{PPDB supervision -- Additive Composition} \\
\tblpad PSL$^\dagger$ \citep{Wieting2016TowardsUP} & 55.07 & 48.00 & 61.63 & 61.21 & 51.32 & 36.29 & 66.52 \\
\tblpad PSL + WR \citep{arora2017simplebut:sif} & 59.5 & 61.8 & 73.5 & 76.3 & 72.0$^\ast$ & 49.0 & \uline{\textbf{72.9}} \\
\tblpad PSL + WR$^\dagger$ \citep{arora2017simplebut:sif} & 64.76 & 62.34 & 73.77 & 73.82 & 70.73 & 45.97 & 70.88 \\
\tblpad PSL + UP \citep{ethayarajh:2018:W18-30:uSIF} & \uline{\textbf{65.8}} & \uline{\textbf{65.2}} & \uline{\textbf{75.9}} & \textbf{77.6} & \uline{\textbf{74.8}} & - & 72.3 \\
\tblpad PSL + UP$^\dagger$ \citep{ethayarajh:2018:W18-30:uSIF} & 65.79 & 64.48 & 75.70 & 76.79 & 74.13 & \textbf{50.64} & 71.80 \\
\multicolumn{1}{l}{PPDB supervision -- Considering Word Alignment} \\
\tblpad WMD\; PSL$^\dagger$ \citep{pmlr-v37-kusnerb15:from} & 55.52 & 44.52 & 61.39 & 69.38 & 56.93 & 50.57 & 61.78 \\
\tblpad WMD\; PSL \small{w/o stopwords}$^\dagger$ \citep{pmlr-v37-kusnerb15:from} & 61.28 & 54.13 & 69.45 & 74.14 & 70.93 & 46.31 & 63.24 \\
\tblpad DynaMax\; PSL \citep{zhelezniak2019:goforthemax:fuzzyjaccard} & 58.2 & 54.3 & 66.2 & 72.4 & - & - & - \\
\tblpad BERTScore\; PSL$^\dagger$ \citep{zhang2019bertscore} & 56.90 & 51.31 & 66.39 & 71.85 & 60.33 & 49.47 & 67.40 \\
\tblpad \textbf{WRD}\; PSL & \textsl{57.84} & \textsl{48.84} & \textsl{63.41} & \textsl{71.20} & \textsl{59.03} & \textsl{48.60} & \textsl{64.29} \\
\tblpad \textbf{WRD}\; PSL + \textbf{VC}(WR) & \textsl{65.13} & \textsl{60.07} & \textsl{71.29} & \textsl{77.20} & \textsl{72.71} & \textsl{52.02} & \textsl{67.44} \\
\tblpad \textbf{WRD}\; PSL + \textbf{VC}(SUP) & \textsl{\textbf{65.60}} & \textsl{\textbf{60.24}} & \textsl{\textbf{72.51}} & \textsl{\uline{\textbf{77.61}}} & \textsl{\textbf{74.31}} & \textsl{\uline{\textbf{54.02}}} & \textsl{\textbf{67.72}} \\
\midrule
\multicolumn{1}{l}{ParaNMT supervision -- Additive Composition} \\
\tblpad ParaNMT$^\dagger$ \citep{wieting2018paranmt} & 67.77 & 62.35 & 77.29 & \textbf{79.51} & \uline{\textbf{79.85}} & \textbf{49.53} & \uline{\textbf{74.80}} \\
\tblpad ParaNMT + WR$^\dagger$ \citep{arora2017simplebut:sif} & 67.81 & 64.62 & 77.00 & 77.87 & 79.74 & 39.42 & 73.48 \\
\tblpad ParaNMT + UP \citep{ethayarajh:2018:W18-30:uSIF} & 68.3 & \uline{\textbf{66.1}} & \uline{\textbf{78.4}} & 79.0 & 79.5 & - & 73.5 \\
\tblpad ParaNMT + UP$^\dagger$ \citep{ethayarajh:2018:W18-30:uSIF} & \uline{\textbf{68.47}} & 65.29 & 78.29 & 78.95 & 79.43 & 46.67 & 73.28 \\
\multicolumn{1}{l}{ParaNMT supervision -- Considering Word Alignment} \\
\tblpad WMD\; ParaNMT$^\dagger$ \citep{pmlr-v37-kusnerb15:from} & 60.06 & 47.00 & 64.01 & 70.40 & 56.43 & 46.95 & 65.06 \\
\tblpad WMD\; ParaNMT \small{w/o stopwords}$^\dagger$ \citep{pmlr-v37-kusnerb15:from} & 63.02 & 54.39 & 70.70 & 73.88 & 72.65 & 47.14 & 64.80 \\
\tblpad DynaMax\; ParaNMT \citep{zhelezniak2019:goforthemax:fuzzyjaccard} & 66.0 & \textbf{65.7} & \textbf{75.9} & \uline{\textbf{80.1}} & - & - & - \\
\tblpad BERTScore\; ParaNMT$^\dagger$ \citep{zhang2019bertscore} & 57.41 & 49.35 & 65.88 & 71.66 & 61.24 & \uline{\textbf{55.44}} & 67.23 \\
\tblpad \textbf{WRD}\; ParaNMT & \textsl{65.89} & \textsl{56.05} & \textsl{72.03} & \textsl{78.01} & \textsl{74.12} & \textsl{53.83} & \textsl{69.49} \\
\tblpad \textbf{WRD}\; ParaNMT + \textbf{VC}(WR) & \textsl{\textbf{67.95}} & \textsl{61.94} & \textsl{75.70} & \textsl{79.96} & \textsl{79.01} & \textsl{50.19} & \textsl{\textbf{70.42}} \\
\tblpad \textbf{WRD}\; ParaNMT + \textbf{VC}(SUP) & \textsl{67.68} & \textsl{61.98} & \textsl{75.57} & \textsl{79.94} & \textsl{\textbf{79.06}} & \textsl{52.44} & \textsl{69.70} \\
\midrule
\multicolumn{1}{l}{SNLI supervision} \\
\tblpad USE (Transformer)$^\ddagger$ \citep{cer2018use} & 61 & 64 & 71 & 74 & - & - & - \\
\tblpad InferSent$^\ddagger$ \citep{conneau2017infersent} & 61 & 56 & 68 & 71 & 75.8$^\ast$ & - & - \\
\tblpad GenSen {\scriptsize (+STN +Fr +De +NLI +2L +STP)} \citep{subramanian2018gensen} & - & - & - & - & 79.2 & - & 88.8 \\
\midrule \midrule
\multicolumn{1}{l}{\textbf{Supervised}} \\
\tblpad XLNet-large (ensemble) \citep{yang2019xlnet} & - & - & - & - & 93.0 & - & - \\
\bottomrule
\end{tabular}
\egroup

\caption{%
    Pearson's $r \times 100$ between the predicted and gold scores is show.
    The best results in each dataset, word vector, and strategy for computing textual similarity (``Additive composition'' or ``Considering Word Alignment'') is in \textbf{bold};
    and the best results regardless of the strategy for computing textual similarity are further \uline{\textbf{underlined}}.
    The results of our methods are \textsl{slanted}.
    Each row marked ($\dagger$) was re-implemented by us.
    Each value marked ($\ddagger$) was taken from \citet{perone2018:evaluationofsentenceembeddings}.
    Each value marked ($\ast$) was taken from STS Wiki (\url{http://ixa2.si.ehu.es/stswiki/index.php/STSbenchmark}).
}
\label{tab:benchmark3}
\end{table*}

\end{document}